\documentclass[%
aps,prd,showpacs,preprintnumbers,amsmath,amssymb,floatfix
]{revtex4-2}

\usepackage{graphicx}
\usepackage{dcolumn}
\usepackage{bm}
\usepackage{lineno}
\usepackage{mathtools}
\usepackage{caption}
\usepackage[subrefformat=parens]{subcaption}
\newcommand{\brain}{``brain"~}

\newcommand{\body}{``body"~}

\newcommand{\wind}{``wind"~}
\newcommand{\windcom}{``wind,"~}
\newcommand{\windper}{``wind."~}

\newcommand{\windpar}{``wind")}

\newif\ifcomment

\begin{document}

\title{Backpropagation through Soft Body: 
\\Investigating Information Processing in Brain--Body Coupling Systems}

\author{Hiroki Tomioka}
\email{tomioka@isi.imi.i.u-tokyo.ac.jp}
\altaffiliation[Current affiliation: ]{Mitsubishi Electric Corporation, Japan. }
\author{Katsuma Inoue}
\affiliation{Graduate School of Information Science and Technology, The University of Tokyo, Tokyo 113-8656, Japan}
\author{Yasuo Kuniyoshi}
\affiliation{Graduate School of Information Science and Technology, The University of Tokyo, Tokyo 113-8656, Japan}
\author{Kohei Nakajima}
\affiliation{Graduate School of Information Science and Technology, The University of Tokyo, Tokyo 113-8656, Japan}




\begin{abstract}
Animals achieve sophisticated behavioral control through dynamic coupling of the brain, body, and environment.
Accordingly, the co-design approach, in which both the controllers and the physical properties are optimized simultaneously, has been suggested for generating refined agents without designing each component separately.
In this study, we aim to reveal how the function of the information processing is distributed between brains and bodies while applying the co-design approach.
Using a framework called ``backpropagation through soft body," we developed agents to perform specified tasks and analyzed their mechanisms.
The tasks included classification and corresponding behavioral association, nonlinear dynamical system emulation, and autonomous behavioral generation.
In each case, our analyses revealed reciprocal relationships between the brains and bodies.
In addition, we show that optimized brain functionalities can be embedded into bodies using physical reservoir computing techniques.
Our results pave the way for efficient designs of brain--body coupling systems.
The source code and supplementary videos are available at: \url{https://github.com/hiroki-tomioka/BPTSB}.
\end{abstract}

\maketitle


\section{Introduction}
In living organisms, the brain and the body do not usually operate separately, which makes the organisms flexible and robust against drastic environmental changes.
Inspired by this relation, morphological computation~\cite{paul2006morphological,pfeifer2006body,pfeifer2009morphological}, which is a research field that utilizes physical characteristics inherent in bodies for information processing, has been developed.
Passive dynamic walkers~\cite{collins2001three}, which achieve smooth bipedal walking with no controller, and ``Dead trout"~\cite{liao2004neuromuscular}, which move upstream in the water by exploiting the interaction between Karman vortexes and their soft bodies, are typical examples.
These studies emphasize that skillful motions do not necessarily require top--down control by the brain. 
Rather they underscore the importance of the interactions between the body and its environment.
This idea that the relationship between the brain, body, and environment produces valuable information is called embodiment~\cite{pfeifer2006body,pfeifer2007self,pfeifer2001understanding,suzumori2023science}.

Recently, a type of physical neural network (NN) called a mechanical NN has been developed.
In this model, an artificial NN is emulated in a mechanical realm using voice coil actuators~\cite{lee2022mechanical}.
In physical reservoir computing~\cite{nakajima2020physical,nakajima2021reservoir}, one of the major morphological computation approaches, information is processed based on the dynamics of the actual physical system.
Examples include water surface~\cite{fernando2003pattern} and fluid dynamics~\cite{goto2021twin}, an octopus-inspired soft silicone arm~\cite{nakajima2013soft,nakajima2015information,nakajima2014exploiting,nakajima2018exploiting,hasegawa2024takorobo}, a quadrupedal robot with flexible spine~\cite{zhao2013spine}, an artificial muscle~\cite{akashi2024embedding,eder2018morphological,hayashi2022online,sakurai2022durable,shinkawa2023limit,kawase2021pneumatic}, a swimming robot~\cite{horii2021physical,he2025physical}, tensegrity~\cite{caluwaerts2014design,caluwaerts2013locomotion}, an origami robot~\cite{bhovad2021physical,wang2023building}, and wings on a flying robot~\cite{tanaka2021flapping}.
In this way, many studies have already proven that the brain is not the only system that processes information.
Morphological computation suggests that the brain, body, and environment are intertwined and should not be considered in an isolated manner.

In traditional robotics, the controllers and mechanical bodies are designed separately.
In typical cases, the controller, which is composed of some algorithms or NNs, is responsible for processing inputs to generate control signals.
Subsequently, the mechanical body executes the command and takes action against the external world.
However, if this design philosophy is followed, the interactions between the controller and the body cannot be fully utilized.
The concept of training a \brain and \body simultaneously, which is called co-designs~\cite{sims1994evolving}, has been proposed from this perspective.
In this study, the module that processes external signals or generates control signals is referred to as the \brain to distinguish it from the \body that represents the physical system.

\begin{figure*}[t]
  \centering
  \includegraphics[keepaspectratio,width=\linewidth,clip]{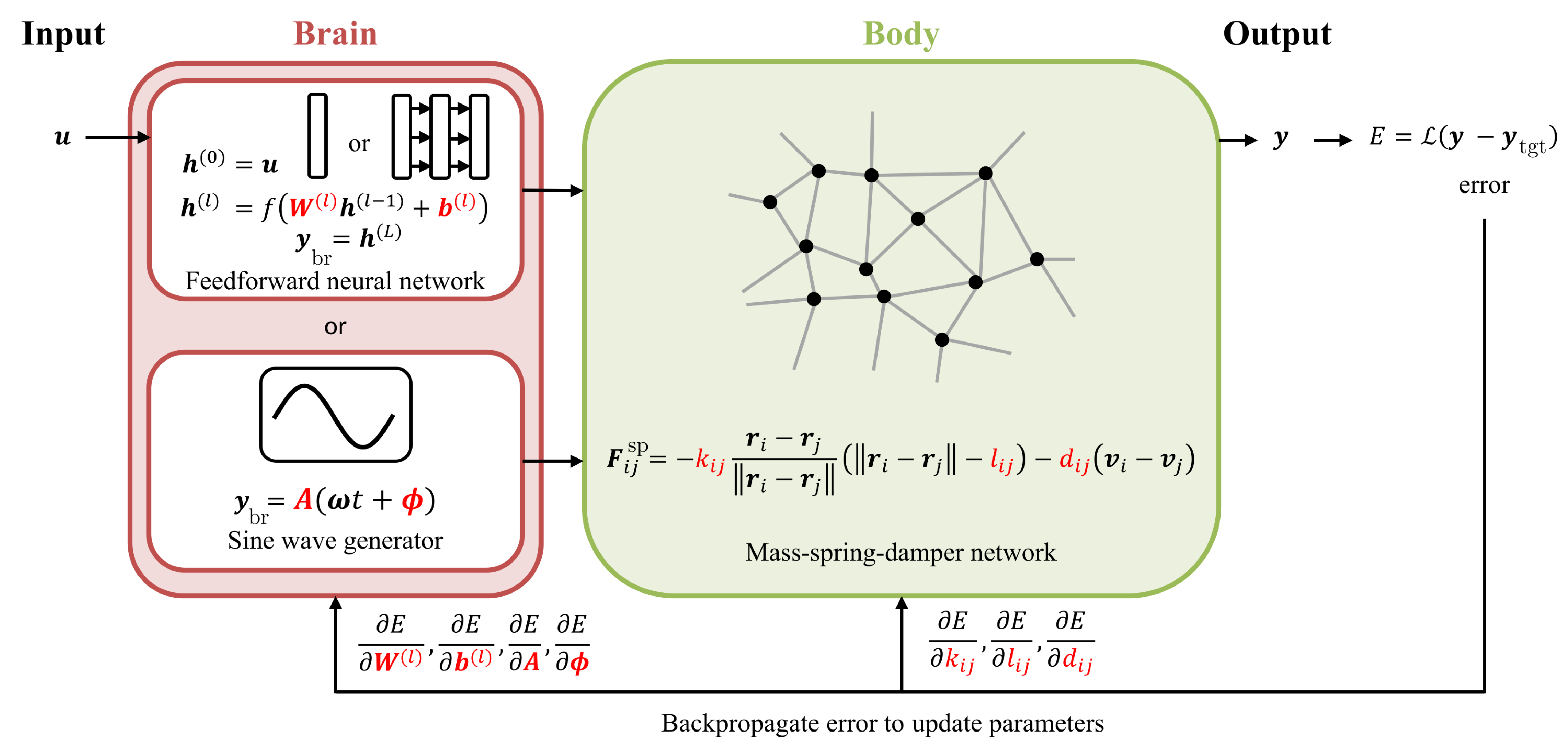}
  \caption{Backpropagation through soft body.
  The system is composed of a brain and a body.
  The brain is a feed-forward neural network (FNN) or a sine wave generator (SWG), and its output $\bm{y}_{\text{br}}$ corresponds to the input to the body.
  The body, a mass-spring-damper network (MSDN), produces physical movements through spring and external forces.
  The input $\bm{u}$ is conveyed to the brain, and the output $\bm{y}$ is generated via information processing and dynamics.
  While calculating the error $E$ with a loss function $\mathcal{L}$ based on the generated output $\bm{y}$ and the target output $\bm{y}_{\text{tgt}}$, the parameters are updated by propagating errors backward through the entire system.
  The parameters in red (the weights of the FNN $W$ and $b$, the amplitude $A$ and phase $\phi$ of the SWG, and the spring constant $k$, damping coefficient $d$, and rest length $l$ of the springs) are trainable.}
  \label{fig:BPTSB}
\end{figure*}

In recent years, soft robotics has become an active research topic, and evolutionary algorithms and differentiable physics simulations have been investigated~\cite{cheney2018scalable}.
In work on differentiable physics simulations, the concept of co-design has emerged in many studies~\cite{du2021underwater,hu2019chainqueen,ma2021diffaqua,wang2023softzoo,bielawski2024best} starting with the research by Hermans et al.~\cite{hermans2014automated}, which updated physical and control parameters using gradient descent.
Thus far, the main theme of the co-design research field is the optimization of a brain--body coupling agent with good performance.
In the current study, we also use differentiable physics simulations with gradient descent optimizations.
But we aim to achieve the following novel objectives: (1) to quantitatively clarify the functional division of roles between the brain and body, which is generated when task-solving ability is acquired as a whole system, and (2) to achieve closed-loop control by embedding a part of the acquired function of the brain into the body.
The method proposed in our current study accomplishes closed-loop control without a target output for feedback from the body in advance, which provides an interesting new perspective in the context of physical reservoir computing.

In this study, a gradient-descent-based method was applied to simultaneously train a brain and a body for co-design, which we call ``backpropagation through soft body" (Fig.~\ref{fig:BPTSB}).
Here, a mass-spring-damper (MSD) system represents the soft body.
We analyzed the functional differentiation and division of roles obtained through optimization for action generation based on image classification or the prediction of nonlinear dynamics, revealing the concealed structure of information processing underlying brain--body coupling systems.
Furthermore, we achieved autonomous behavioral generation based on the controls for cyclic motion generation and locomotion.
By embedding a portion of the obtained brain functionality into the body, closed-loop control was achieved, which facilitates the maximally effective use of body dynamics.

\section{Results}
\subsection{Body model and brain model}
In this study, we adopted an MSD network (MSDN) as a model of a mechanical system (Fig.~\ref{fig:BPTSB}).
MSD systems are frequently used as models of soft bodies~\cite{hauser2011towards,hauser2012role,haghshenas2022exploiting,hauser2011moving,hermans2014automated,urbain2017morphological,komatsu2021algebraic}.
In our experiments, the mass points to be fixed at coordinates are called fixed mass points, and those that can move under the action of the force from springs or external forces are movable mass points.
These numbers are described as $N_{\text{fix}}$ and $N_{\text{mov}}$, respectively.
Our simulation environment was a two-dimensional plane, and the components of the MSD system are simply called springs.
Each spring is linear; however, the whole body system exhibits nonlinearity because of the geometric constraints of MSDNs (Section~\ref{sec:MSDN_nonlinearity}).

We define the brain as a structure separate from the body.
Together, the brain and body constitute a system that processes information (Fig.~\ref{fig:BPTSB}). 
In our experiments, the brain had two functions: (1) the transformation of inputs and (2) the generation of periodic signals.
The former delivers the converted external information to the body.
In particular, we implemented feed-forward NNs (FNNs): either a single linear input layer (LIL) or a multilayer perceptron (MLP).
While the LIL performed simple linear transformations, the MLP incorporated nonlinear transformations, thanks to the nonlinear activation function (ReLU).
The latter brain is defined as a sine wave generator (SWG).
The detailed settings of the SWG are presented in Section~\ref{sec:CL}.

\subsection{Backpropagation through soft body}

We built an experimental system for the MSDN simulation while retaining gradients consistently throughout the whole brain--body system.
As for the system that coupled the brain and body, the training proceeded by propagating errors backward through a physical simulation, so this framework is called backpropagation through soft body (BPTSB; Fig.~\ref{fig:BPTSB}) in this paper.
The brain parameters (weight matrix $W$ and $b$, amplitude $A$, and phase $\phi$) and the body parameters (spring constant $k$, damping coefficient $d$, and length $l$) could be trained.
These parameters were updated by propagating errors through the chain rule (see Section~\ref{sec:BPTSB_method}).

\begin{figure*}[t]
  \centering
  \includegraphics[keepaspectratio,width=\linewidth,clip]{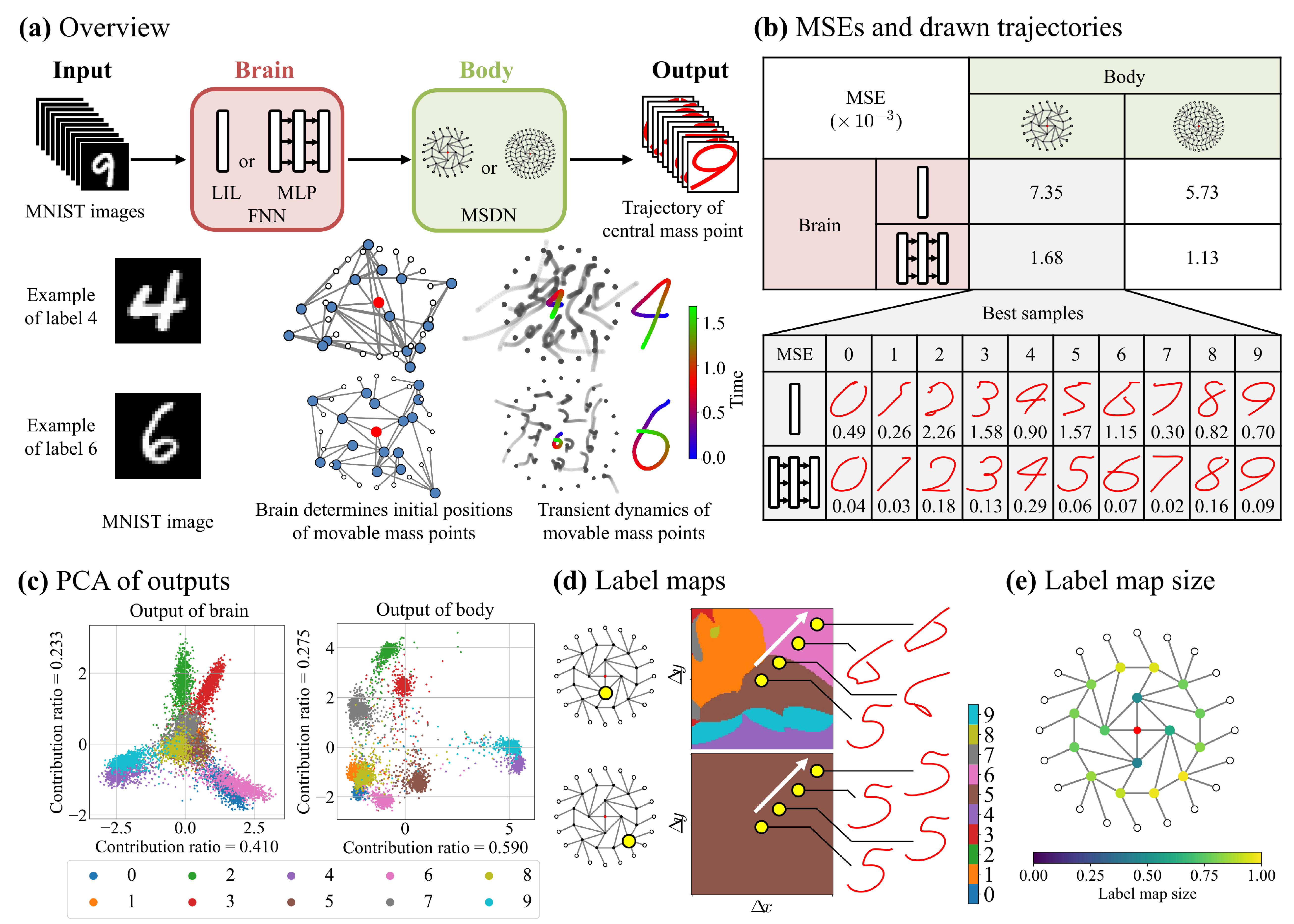}
  \caption{An MNIST classification and drawing task.
    (a) The input, an MNIST image, is conveyed to the brain, and the brain transforms the input information into the initial positions of the movable mass points (blue points; exclude the central mass point (CMP)).
    The trajectory of the CMP becomes the output.
    The trajectory is color-matched with the elapsed time.
    The optimized system draws the shape of the input label number (see Supplementary Video 1).
    (b) The errors calculated based on the drawn trajectories of the trained system with each configuration.
    The table below shows the actual trajectories and the errors of the best samples in their interactions with a multiple-circle body ($N_{\text{mov}}=17$).
    (c) The outputs of the brain (the initial positions of the movable mass points) and the body (the trajectory of the CMP) are compressed into two dimensions using principal component analysis (PCA).
    Each color corresponds to the input label.
    The more separated the dots on the basis of color, the stronger the label classification.
    (d) Color coding is performed based on the trajectory drawn via transient dynamics.
    Among the initial positions of the movable mass points, which are determined by the brain, one mass point (yellow) is displaced, and the dynamics are then measured. 
    The colors correspond to the labels with the least error between the actual and target trajectories.
    For instance, the brown area indicates that the shape of the trajectory is most similar to a five.
    Two examples in which one mass point is displaced from its initial position for the label ``five" are displayed.
    As the point is shifted to the upper right, the trajectory changes to six in the top figure.
    However, in the lower part of the figure, the trajectory is almost constant.
    (e) The color represents the degree of simplicity of the label maps, which means that the trajectory becomes more sensitive to the initial position as the value decreases.}
  \label{fig:MNIST}
\end{figure*}

\subsection{Division of functional roles}
\label{sec:MNIST}

In this section, we investigate the division of functional roles between the brain and body resulting from simultaneous optimization.
Using the Modified National Institute of Standards and Technology database (MNIST)~\cite{lecun1998gradient}, handwritten characters were classified by the brain--body coupling system.

In this task, taking an LIL or an MLP as the brain, we adopted a multiple-circled-shaped ($N_\text{mov}=17,65$) MSDN as a body.
The system input was an image from the MNIST, and the system output was the trajectory of the central mass point (CMP) of the MSDN, where the aim was to draw the corresponding number with it (Fig.~\ref{fig:MNIST} (a)).
A $28\times28$-dimensional image was input to the brain, followed by a transformation into the initial positions of the movable mass points (excluding the CMP; $2(N_{\text{mov}}-1)$-dimension).
The mass points were released from their initial positions at rest at time $t=0$, and motions were then generated through transient dynamics without external forces.
Regarding the trajectory of the CMP as the output, the system was trained to draw the shape of the input label's number.
For example, when the input image had the label ``four," the CMP was expected to move in the shape of a four.
The functionalities required for this task can be divided into two types: recognition and control.
Recognition is defined as the classification of input images, and control corresponds to the subsequent drawing behavior.
Accordingly, in a top-down manner, we would separately design a brain to produce unique outputs according to the input labels, and a body to draw 10 trajectories from zero to nine, depending on the 10 types of initial positions.
Furthermore, we need to connect them appropriately in series.
The aim of this experiment was to investigate whether the system would be optimized with a complete functional division of roles or as an intertwined system whose functions are not separated.

The results in Fig.~\ref{fig:MNIST} (b) show cases in which four systems combining two types of brains and bodies were optimized.
The actual drawing behaviors can be seen in Supplementary Video 1.
As for the brain, the error was smaller in the MLP than in the LIL, reflecting the contribution of the nonlinearity of the brain to its performance.
Focusing on the body, increasing the mass points led to better performance, suggesting an increase in the expressiveness of the physical system.

We investigated whether the functionalities of recognition and control were divided between the brain and the body.
To determine this division of roles, we focused on the outputs of each component.
The former corresponds to the initial positions of the mass points, and the latter corresponds to the trajectory of the CMP.
Supposing that the brain is responsible for the classification functionality, the degree of separation of the output from the brain and that of the body should be about the same.
We verified this by compressing these outputs into two dimensions using principal component analysis (PCA) (Fig.~\ref{fig:MNIST} (c)).
Although each output classified the input labels, more separation was emphasized, especially with respect to labels 5 (brown), 7 (gray), and 8 (yellow) in the output of the body.
This suggests that the body enhanced the classification performance in addition to the information processing in the brain.
When the input was limited to a specific set of images, the improvement in separation performance via the body was more pronounced (see Section~\ref{sec:specific_MNIST}).

When then investigated the body's responsiveness to the initial positions to clarify the roles of the brain and the body.
We analyzed the effect on the trajectory of the CMP of changes in the initial positions of the noncentral mass points.
Diagrams were created in which the displacements of the noncentral mass points were charted on $x$-$y$ axes (Fig.~\ref{fig:MNIST} (d)).
These diagrams are called label maps, and their colors correspond to the numbers to which the trajectory shapes of the CMP were most similar.
The detailed procedures are presented in Section~\ref{sec:MNIST_method}.
The map structures of the mass points directly connected to the CMP were intricate, suggesting that the trajectories tended to change qualitatively; as the mass point was shifted, for example, the trajectory changed from a ``five" shape to ``six," as shown at the top of Fig.~\ref{fig:MNIST} (d).
Conversely, other maps were mostly occupied by the original label number (see Section~\ref{sec:label_maps}).

In label maps, the ability to follow a trajectory close to the original label number despite changes in the initial position is tied to the robustness of the mass point.
Here, the proportion of the original label area is called the label map size, which quantitatively indicates the robustness of each movable mass point with respect to its initial position (Fig.~\ref{fig:MNIST} (e)).
The smaller the label map size, the more sensitive the mass point, which signifies low robustness.
Therefore, the classification capability of the brain is important.
Conversely, when a label map is simple, even if the classification performed by the brain does not reflect high accuracy, it is possible that the body has acquired a function to absorb the differences.
To shed light on this distribution, the label map size of the mass point that connects directly to the CMP is smaller.
Since it has a spring that can apply force directly to the CMP, it is assumed to be more sensitive to positional changes.

\begin{figure*}[t]
  \centering
  \includegraphics[keepaspectratio,width=\linewidth,clip]{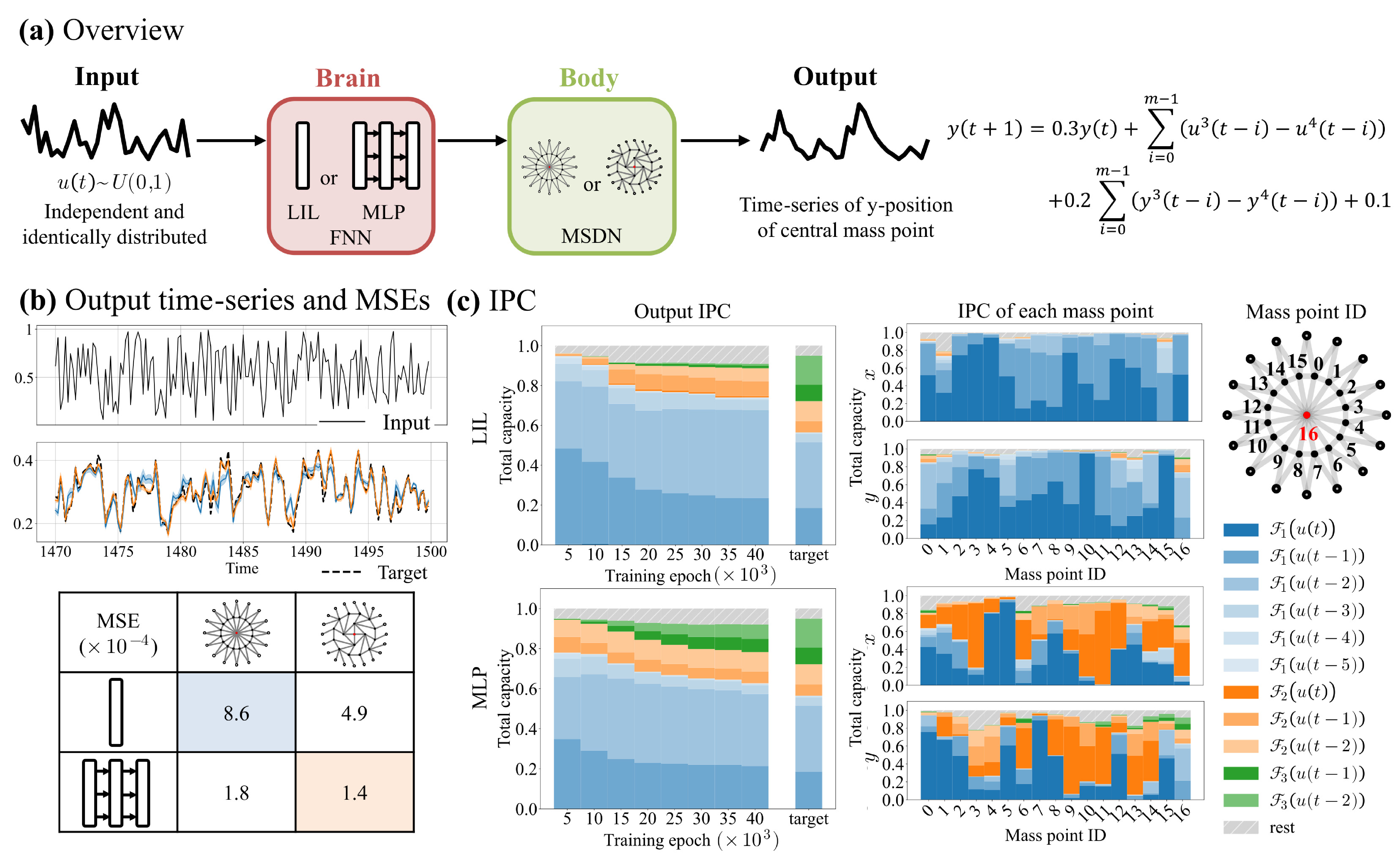}
  \caption{Time-series emulation task.
    (a) The input is an independent and identically distributed uniform random value, and the output corresponds to the $y$-axis position of the CMP.
    The target output is calculated via the nonlinear transformation of the past input and output values.
    The one-dimensional input $u(t)$ is transformed by the brain into external forces and then sequentially conveyed to the movable mass points (except the CMP).
    While the body behaves in accordance with the external forces, the $y$-coordinate of the CMP corresponds to the system output $y(t+1)$ (Supplementary Video 2).
    (b) The input, target, and output time-series.
    The table below lists the reconstruction errors of each system, with the background color corresponding to the color of the time-series plots.
    (c) The left side represents the information processing capacity (IPC) values of the system output, which are arranged from left to right according to training epochs.
    The rightmost bar represents the IPC value of the target time-series.
    The middle figures represent the IPC of each movable mass point (one dimension each in the $x$ and $y$ axes).
    These are the results of the system that combines a double-circle structured body with an LIL or an MLP, where the mass point ID is given in a clockwise direction, as shown in the upper right.
    The degrees of the orthogonal polynomials and the delayed time steps of the input are given as $n$ and $\Delta t$, respectively, in $\mathcal{F}_n\left(u(t-\Delta t)\right)$, and they are equivalent to the colors and intensity of the bars.
    In particular, the blue parts correspond to linear profiles.
    The total capacity reached the rank of one in all cases.}
    \label{fig:time-series}
\end{figure*}

\subsection{Distribution of computational properties}
\label{sec:timeseries}

To investigate functionality in greater detail, we examined the distribution of computational properties, especially nonlinearity.
We worked through a time-series emulation task in which a system receives input time-series and generates output time-series to emulate the target.
The system is required to reconstruct the target time-series based on the received input (Fig.~\ref{fig:time-series} (a)).
With reference to the nonlinear autoregressive moving average (NARMA), we prepared the target output time-series:
\begin{align}
  y(t+1)=&0.3y(t)+\sum_{i=0}^{m-1}\left(u(t-i)^3-u(t-i)^4\right)+0.2\sum_{i=0}^{m-1}\left(y(t-i)^3-y(t-i)^4\right)+0.1.
\end{align}

An LIL and an MLP were used as brains to verify the contribution of nonlinearity to the task performance.
The body had double-circle ($N_\text{mov}=17$) or multiple-circle ($N_\text{mov}=17$) structures.
The input was converted to $2(N_{\text{mov}}-1)$-dimensional external forces in the brain to act on the movable mass points, except for the CMP.
The body generated the dynamics, and the $y$-coordinate position of the CMP was designated as the one-dimensional output.

As for the brain, while the structure of the target output time-series could be captured to some extent with the LIL, the MLP yielded higher reconstruction accuracy (Fig.~\ref{fig:time-series} (b)).
As we found when comparing the bodies, there was a tendency for the error to be smaller in the multiple-circle structure.
This trend reflects the contribution of the diversity of the geometric structures of the body to the nonlinearity.

To analyze the systems, the information processing capacity (IPC)~\cite{dambre2012information} was used to clarify the profile in terms of memory and nonlinearity (Fig.~\ref{fig:time-series} (c)).
The system output became asymptotic to the target model as training progressed.
Not only the first-order profile (linear) but also the second- and third-order profiles increased, suggesting that the optimization proceeded to enhance higher-order nonlinearity.
However, because the nonlinearity in the system with the LIL depended only on the geometric conditions of the body, the third-order profile was too small for reconstruction.
On the other hand, the MLP could apply nonlinear transformation as well as the body, so the system with the MLP comparatively succeeded in reproducing the target nonlinearity.
In comparison with the LIL, the MLP system demonstrated higher-order nonlinearity at each mass point as well.
This is presumably because the external force input to the body was uniformly transformed by the ReLU function in the MLP.
The memory (the delayed time step of the input) was generated by the body since the brain was an FNN.
Hence, it must be true that the nonlinear information that the brain delivered to the body was transformed into nonlinear memories through the dynamics of the body.
Please refer to Section~\ref{sec:memory_task} for the experiments and further analysis of memory.

\begin{figure*}[t]
  \centering
  \includegraphics[keepaspectratio,width=\linewidth,clip]{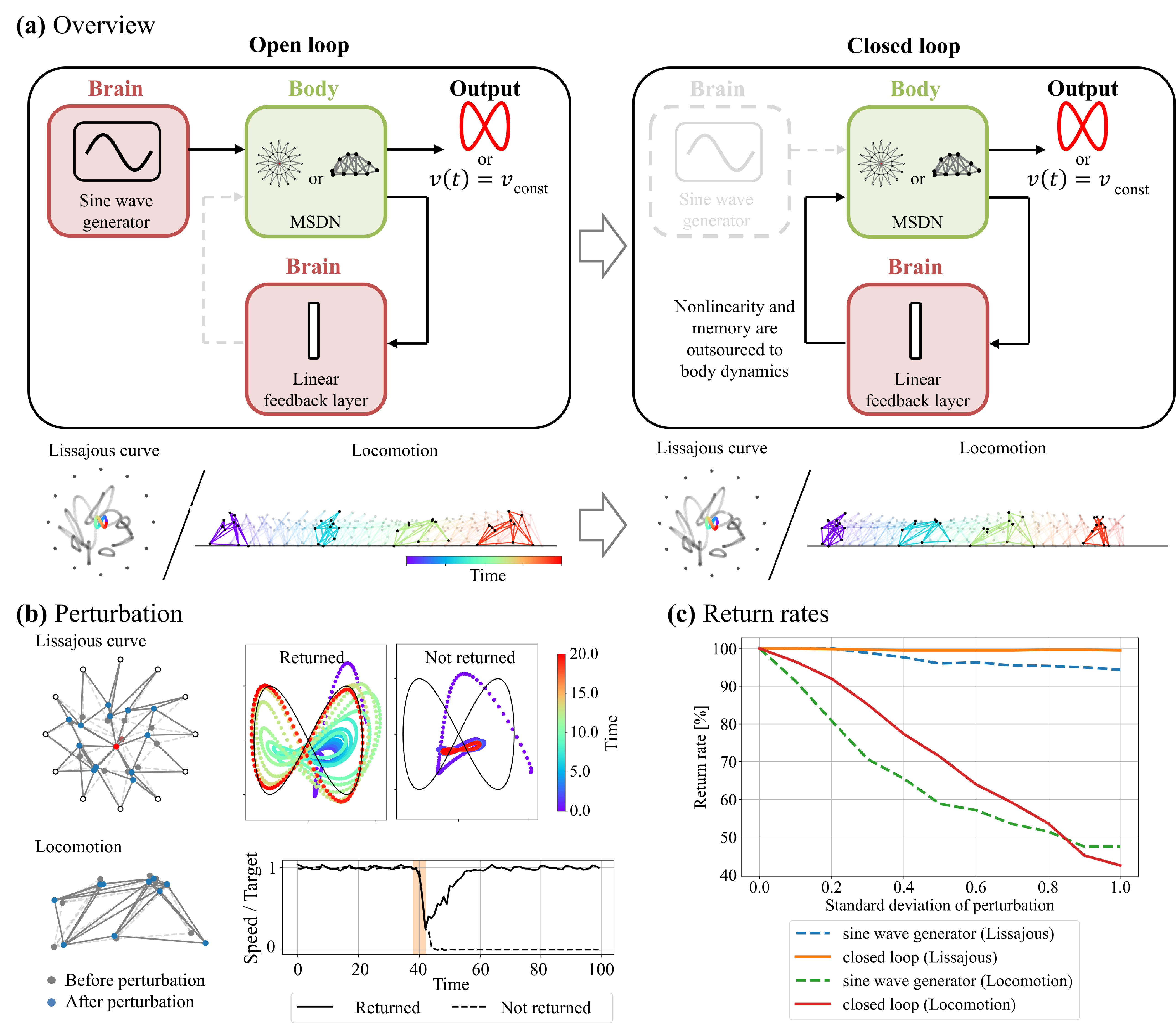}
  \caption{Closed-loop control by feedback layer.
    (a) A feedback layer (FL) is added to an open-loop system composed of an SWG and an MSDN, and it is then trained in an open loop.
    After this training, a closed loop is formed by designating the FL as the brain instead of the SWG.
    The lower row shows the behaviors in an open loop and a closed loop (Supplementary Video 3 and 4).
    (b) The robustness is tested by applying perturbations to the positions of the movable mass points (shifting the gray points to the blue points) and checking if the system returns to the desired behavior afterwards.
    In the Lissajous curve drawing task, the upper right figures show the trajectories of the systems; one could return and the other could not return to the trained trajectory.
    In the locomotion task, the locomotion speeds of the systems are displayed in the lower right.
    A perturbation was applied to the system according to the timing of the orange background color.
    (c) It shows the rate at which the system returns to the trained behavior based on the magnitude of the perturbation.
    The standard deviation when generating a perturbation is parameterized, which corresponds to the value of the horizontal axis.}
    \label{fig:closed-loop}
\end{figure*}

\subsection{Autonomous behavior through brain function transfer}
\label{sec:CL}

In the work reported in this section, we aimed to realize autonomous behaviors by designing the functional transfer of a brain after the simultaneous optimization of brain--body coupling systems.
We verified the division of roles between a brain and a body by embedding the trained command into a closed-loop system.
The framework (Fig.~\ref{fig:closed-loop} (a)) is as follows:
\begin{enumerate}
  \item A system composed of an SWG and an MSDN is trained for an objective behavior by BPTSB.
  \item The weights of a feedback layer (FL) are calculated via ridge regression based on the error between the output of the SWG and that of the FL.
  \item A closed loop is formed by replacing control signals with feedback signals.
\end{enumerate}
An SWG has trainable parameters, namely, an amplitude $A_{ij}$ and phase $\phi_{ij}$, and the signals generated by the brain modulate the rest length of each spring to drive a body~\cite{urbain2017morphological}.
In Section~\ref{sec:switching} and \ref{sec:switching_timing}, we investigate open-loop systems with SWGs.
Utilizing the spring lengths as feedback information, the new brain, an FL, transforms them into control signals.
The advantage of this training framework is that target feedback output is not explicitly required from the beginning.
This is because this method starts with searching for the optimal target output by training the control parameters of the SWG, which is different from normal reservoir computing.

First, the system, whose body was a double-circle MSDN ($N_\text{mov}=13$), was trained so that the CMP drew a Lissajous curve; this task embeds a limit cycle into the brain--body coupling system.
We optimized the system with an SWG and then established a closed loop.
As a result, although the error was larger than it was with the SWG, the system autonomously drew a Lissajous curve by properly feeding back its body dynamics (Fig.~\ref{fig:closed-loop} (a)).

Using similar settings, we can demonstrate locomotion in the context of robotics.
A simple caterpillar-shaped MSDN with 10 mass points and 24 springs was used as a body.
The robot, whose brain was an SWG, was trained to move forward using BPTSB.
Optimization of multiple systems initialized with different random seeds separately led to qualitatively diverse locomotion styles.
For instance, there was a system that ran in a gallop with structures that looked like front and hind legs, and there was another system that moved forward while repeatedly jumping by expanding and contracting up and down (Supplementary Video 4).
An FL was subsequently trained.
Here, success in autonomous locomotion was declared when the ratios of the locomotion speed in closed-loop control and in open-loop control were 0.7 or more.
When we created closed-loop systems using 20 agents that learned forward behavior from random initialization, 8 agents (40\%) succeeded.

We then analyzed the stability of the closed-loop systems.
After applying perturbations to the positions of the movable mass points, we checked whether the system returned to the desired behavior (Fig.~\ref{fig:closed-loop} (b)).
A noise with a normal distribution (mean zero) was provided as the perturbation.
The perturbation was parameterized based on the standard deviation of the noise distribution.
As a measure for evaluating robustness, the ``successful return" is described for each task in Section~\ref{sec:Lissajous_method} and~\ref{sec:locomotion_method}.
In the Lissajous task, while there were systems that successfully returned to the trained trajectory, systems that did not recover finally converged to untrained attractors (Fig.~\ref{fig:closed-loop} (b)).
In the locomotion task, cases successfully returned by gradually regaining their speed.
On the other hand, there were some agents that suddenly slowed down and came to a standstill.
As the standard deviation of the perturbation increased, the return rates tended to decrease (Fig.~\ref{fig:closed-loop} (c)).
However, the return rates of the sinusoidal drive and closed-loop control were almost the same, suggesting that the robustness was not impaired by the closed loop.

\section{Discussion}
In this study, we applied BPTSB to elucidate the underlying structures of brain--body coupling systems.
This framework is similar to the corresponding methods in the fields of differentiable physics simulation~\cite{du2021underwater,hermans2014automated,hu2019chainqueen,ma2021diffaqua,wang2023softzoo} and reservoir computing~\cite{subramoney2021reservoirs}.

In the MNIST task, we investigated whether two functions, recognition and control, were shared by the brain and body.
It became clear that label classification was not completed solely in the brain.
The body was trained to absorb the differences in the output from the brain, leading to an almost constant trajectory, depending on the input label.
In short, the roles that emerged from simultaneous optimization were not completely divided into top--down recognition and control, but were rather optimized within a system that performed a task complementary to each other.

We then worked on a time-series emulation task and analyzed the optimized systems using IPC.
In this experiment, it was shown that higher-order nonlinearity that was insufficient in the body alone was compensated for by the MLP, and this satisfied the information processing requirements on the whole.
On the contrary, due to the lack of memories in the brain (FNN), the body elevated the nonlinearity, which was conveyed from the brain, into nonlinear memories.

These two experiments demonstrate that the functionality of a brain--body system as a whole is realized through the reciprocal relationship between the brain and the body.
By designing and optimizing brains and bodies as joint systems, we can obtain information processing structures that are not possible if they are designed separately.

Based on this knowledge, we propose a method for constructing a closed-loop control without target feedback signals beforehand.
In particular, after optimizing a brain (SWG) and a body for an objective behavior, closed-loop control was accomplished by transferring the control signal generation to an FL.
A similar concept has been proposed by Urbain et al.~\cite{urbain2017morphological}.
Lissajous curve drawing and locomotion tasks showed that it is possible to establish a closed-loop system that generates autonomous behaviors through feedback between its own physical states.
Responses to perturbations demonstrated the robustness of the closed-loop system.
In the locomotion task, although the closed-loop control was successful in 40\% of all systems, a higher rate may be achievable through the modification of the body states or the implementation of a transitional period between the open and closed loops~\cite{urbain2017morphological}.
Moreover, it may be possible to optimize a brain and a body for a closed loop simultaneously (Section~\ref{sec:CL_BPTSB}).
Although the validity of the performance level examined in this study was unclear, the concept of optimizing a sensorimotor coupling system to embed autonomous controls through simultaneous training is a novel proposal in this field.

We used BPTSB, a gradient-based method, but this is just one example of an optimization method.
Methods of optimizing systems are divided into two types based on whether they are gradient-based.
Gradient-based methods can perform training more rapidly and efficiently.
In addition, it is possible to reflect the physical properties of the system~\cite{pinskier2022bioinspiration}.
Researchers have examined systems that integrate actual physics with schemes for gradient calculation~\cite{hermans2015trainable,wright2022deep}, especially in optical NNs~\cite{wagner1987multilayer,wagner1993optical,hughes2018training,guo2021backpropagation}.
In recent years, the main approach to gradient-based methods is differentiable physics simulation (rigid bodies~\cite{de2018end,degrave2019differentiable,freeman2021brax} and soft objects~\cite{du2021diffpd,du2020functional,heiden2021neuralsim,hu2019difftaichi,hu2019chainqueen,qiao2021differentiable}).
On the other hand, non-gradient-based methods can be applied to systems in the real world, and they also have an advantage in scalability.
One example of such a method is evolutionary robotics~\cite{floreano2008bio,nolfi2000evolutionary}.
Researchers have also trained systems based on their physical properties~\cite{rocks2017designing,stern2020supervised}.
Augmented direct feedback alignment~\cite{nakajima2022physical} is another method that avoids gradient computation.
One possible direction for further developing our research would be to investigate whether adopting these methods results in different functional distributions between brains and bodies.
Our subsequent work will involve actual physics in the real world.

\section{Methods}
\subsection{Body model}
Inspired by a study that made use of geometrically constructed MSDNs~\cite{hermans2014automated}, we incorporated an MSDN into our experiments by performing the physical calculations described below.
The components of the MSDN described in this paper were linear springs, and the force acting on the mass point $i$ was limited to that derived from the connecting springs when there was no external force.
The force exerted on the mass point $i$ by a spring connected to the mass point $j$ is expressed as
\begin{align}
  \bm{F}_{ij}^\text{sp}=-k_{ij}\frac{\bm{r}_i-\bm{r}_j}{\lVert\bm{r}_i-\bm{r}_j\rVert}\left(\lVert\bm{r}_i-\bm{r}_j\rVert-l_{ij}\right)-d_{ij}\left(\bm{v}_i-\bm{v}_j\right),
\end{align}
where $\bm{r}_i$ and $\bm{v}_i$ are the position and velocity of mass point $i$, respectively. The variables $k_{ij}$, $d_{ij}$, and $l_{ij}$ indicate the spring constant, damping coefficient, and rest length of the spring connecting mass points $i$ and $j$, respectively.
Therefore, the force exerted on mass point $i$ is given by
\begin{align}
  \label{eq:force_only_spring}
  \bm{F}_{i}=\sum_j \bm{F}_{ij}^\text{sp}.
\end{align}
In the physical simulation performed in this study, the effects of collisions between the mass points and intersections of the springs were not considered.

There are two main types of body shapes.
One is a circular shape and the other is a caterpillar-shaped robot structure~\cite{urbain2017morphological}.
The mass points that make up the outermost circle of a circular MSDN are always fixed at coordinates (fixed mass points), and the others are assigned to movable mass points.
Among the movable mass points, the one located in the center is called the central mass point (CMP).
All movable mass points of a double-circle structure are connected to the CMP by one spring, whereas those of a multiple-circle structure do not have this limit.
Meanwhile, the caterpillar-shaped network was exploited as a robot agent in locomotion experiments.
All the mass points were movable, namely $N_{\text{fix}}=0$.
It is noteworthy that although the components of the mechanical network were linear springs, the whole network demonstrated nonlinearity due to the geometrical constraints of MSDNs (Section~\ref{sec:MSDN_nonlinearity}). 

The trainable body parameters include only the physical parameters of the springs; therefore, the topology, which includes the connective relation of each spring and the number of mass points, was kept fixed.
The physical parameters included the spring constant $k$, damping coefficient $d$, rest length $l$, and mass $m$.
All mass parameters were fixed at one, and the spring constant, damping coefficient, and rest length could be optimized.
For the spring constant and damping coefficient, initial values were randomly assigned to each spring.
The initial spring constants were set from 1 to 100, and the initial damping coefficients were set from 0 to 10.
The initial rest lengths were determined to align with the network shape.
At this point, no force could be generated between the mass points, because the rest lengths were equivalent to the distances between the mass points.
Each physical parameter was assigned a separate learning rate (Section~\ref{sec:learning_rates}), and its value was set to never be less than zero.

For the physical simulation, the Velocity Verlet algorithm~\cite{verlet1967computer} was applied, which is frequently used for molecular dynamics simulations.
The Velocity Verlet algorithm is depicted as follows:
\begin{align}
  \bm{r}(t+\Delta t)=\bm{r}(t)+\bm{v}(t)\Delta t+\frac{\bm{a}(t)}{2}(\Delta t)^2,\\
  \bm{v}(t+\Delta t)=\bm{v}(t)+\frac{\bm{a}(t+\Delta t)+\bm{a}(t)}{2}\Delta t,
\end{align}
where $\bm{a}(t)$ is acceleration. 
Based on the equation of motion, this variable can be calculated as
\begin{align}
  \bm{a}(t)=\frac{\bm{F}(t)}{m},
\end{align}
where $t$ is time, $\Delta t$ is a time step, and $\bm{F}(t)$ is force.
The time step was set to $\Delta t=0.01$.

\subsection{Brain model}
Two very simple models were selected for the brain.
One was an FNN, which is generally used as a brain in the field of machine learning to transform external inputs.
It was a linear layer or an MLP (three layers) in particular.
The initial weights of the linear layer were sampled from a normal distribution with a mean of zero and a standard deviation of one, while those of the MLP were initialized using Xavier initialization~\cite{glorot2010understanding}.
The number of nodes in the middle layers of the MLP was set to 128.
The other brain was defined as an SWG, which moved the body.
The source from which the body was driven was the rest lengths of the springs, which meant that the modulation of the rest lengths produced motions in the body.
The control signal generated in the brain is expressed as follows~\cite{urbain2017morphological}:
\begin{align}
  l_{ij}^\text{M}(t)=l_{ij}\left(1+A_{ij}\sin(\omega t+\phi_{ij})\right),
\end{align}
where $l_{ij}^\text{M}(t)$ is the modulated rest length, $l_{ij}$ is the original rest length, $A_{ij}$ is the amplitude, $\omega$ is the frequency, $\phi_{ij}$ is the phase, and $t$ is the time.
The physical calculation of the MSDN was conducted based on $l_{ij}^\text{M}(t)$.
Among the sinusoidal parameters, the amplitude $A_{ij}$ and phase $\phi_{ij}$ were assigned to be trainable.
The initial values of the amplitude were set to 0.4 or 0.5 depending on the experiments.
The initial values of the phase were randomly sampled from 0 to $2\pi$, whereas the frequency was fixed at $2\pi$.

\subsection{Backpropagation through soft body}
\label{sec:BPTSB_method}
The system input was set as $\bm{u}$, the brain output as $\bm{y}_{\text{br}}$, the body output (equivalent to the system output) as $\bm{y}_{\text{bo}}$, the target output as $\bm{y}_{\text{tgt}}$, the brain parameters as $\boldsymbol{\theta}_{\text{br}}$, and the body parameters as $\boldsymbol{\theta}_{\text{bo}}$.
In this paper, the weights of the FNN $W$ and $b$, amplitude $A$ and phase $\phi$ of the SWG correspond to the brain parameters, and the spring constant $k$, damping coefficient $d$, and rest length $l$ correspond to the body parameters.
When the conversions in the brain and the body are written as $\bm{F}_{\text{br}}$ and $\bm{F}_{\text{bo}}$, respectively, the relation between them is shown by
\begin{align}
    \bm{y}_{\text{br}}&=\bm{F}_{\text{br}}\left(\bm{u};\boldsymbol{\theta}_{\text{br}}\right), \\
  \bm{y}_{\text{bo}}&=\bm{F}_{\text{bo}}\left(\bm{u},\bm{y}_{\text{br}};\boldsymbol{\theta}_{\text{bo}}\right).
\end{align}
In addition, the error $E$ is computed using a loss function $\mathcal{L}$ as folllows:
\begin{align}
    E=\mathcal{L}\left(\bm{y}_{\text{bo}},\bm{y}_{\text{tgt}}\right).
\end{align}
Based on the error $E$, the gradients for parameters $\boldsymbol{\theta}_{\text{br}}$ and $\boldsymbol{\theta}_{\text{bo}}$ were calculated using the chain rule as follows:
\begin{align}
    \frac{\partial E}{\partial \boldsymbol{\theta}_{\text{bo}}}
    &=\frac{\partial E}{\partial \bm{y}_{\text{bo}}}\frac{\partial \bm{y}_{\text{bo}}}{\partial \boldsymbol{\theta}_{\text{bo}}} \nonumber \\
    &=\frac{\partial \mathcal{L}\left(\bm{y}_{\text{bo}},\bm{y}_{\text{tgt}}\right)}{\partial \bm{y}_{\text{bo}}}\frac{\partial \bm{F}_{\text{bo}}\left(\bm{u},\bm{y}_{\text{br}};\boldsymbol{\theta}_{\text{bo}}\right)}{\partial \boldsymbol{\theta}_{\text{bo}}}, \\
    \frac{\partial E}{\partial \boldsymbol{\theta}_{\text{br}}}
    &=\frac{\partial E}{\partial \bm{y}_{\text{br}}}\frac{\partial \bm{y}_{\text{br}}}{\partial \boldsymbol{\theta}_{\text{br}}} \nonumber \\
    &=\frac{\partial E}{\partial \bm{y}_{\text{bo}}}\frac{\partial \bm{y}_{\text{bo}}}{\partial \bm{y}_{\text{br}}}\frac{\partial \bm{y}_{\text{br}}}{\partial \boldsymbol{\theta}_{\text{br}}} \nonumber \\
    &=\frac{\partial \mathcal{L}\left(\bm{y}_{\text{bo}},\bm{y}_{\text{tgt}}\right)}{\partial \bm{y}_{\text{bo}}}\frac{\partial \bm{F}_{\text{bo}}\left(\bm{u},\bm{y}_{\text{br}};\boldsymbol{\theta}_{\text{bo}}\right)}{\partial \bm{y}_{\text{br}}}\frac{\partial \bm{F}_{\text{br}}\left(\bm{u};\boldsymbol{\theta}_{\text{br}}\right)}{\partial \boldsymbol{\theta}_{\text{br}}}.
\end{align}
Using the gradients, each parameter was updated based on the following formulas:
\begin{align}
    \boldsymbol{\theta}_{\text{bo}}&\leftarrow \boldsymbol{\theta}_{\text{bo}}-\boldsymbol{\eta}_{\text{bo}}\frac{\partial E}{\partial \boldsymbol{\theta}_{\text{bo}}}, \\
    \boldsymbol{\theta}_{\text{br}}&\leftarrow \boldsymbol{\theta}_{\text{br}}-\boldsymbol{\eta}_{\text{br}}\frac{\partial E}{\partial \boldsymbol{\theta}_{\text{br}}}.
\end{align}
The variables $\boldsymbol{\eta}_{\text{bo}}$ and $\boldsymbol{\eta}_{\text{br}}$ are the learning rates for the body and the brain, respectively.
The actual values of these variables are summarized in Section~\ref{sec:learning_rates}.

\subsection{MNIST classification and drawing task}
\label{sec:MNIST_method}
The structures of the body used in this experiment had  multiple-circle structures ($N_\text{mov}=17, 65$).
MNIST images were vectorized by simply reshaping them into one dimension.
The brain computed the initial arrangement of the body.
Since the spring lengths of the MSDN placed in the initial position were not equal to their rest lengths, forces were exerted between each spring.
The body released from the initial position at $t=0$ generated transient dynamics driven by these spring forces.
Although the positions of the movable mass points changed according to the transient dynamics, we focused only on the trajectory of the CMP as the output.
Target trajectories from zero to nine, including stroke orders and times to finish drawing, were set in advance to be unique (i.e., the control of the $z$-axis was known).
To restrict the starting position for the drawing, the initial position of the CMP was kept fixed.
Generally, no matter what input image the system received, the position of the CMP at the beginning of the simulation was constant.
However, because the other movable mass points reflected the information in the input image, they affected the CMP, resulting in a variety of trajectories.
The loss function was used to compute the mean squared error between the actual and target trajectories.
The results of the experiments concerned with MNIST classification by reading out the states of the body, which means that the physical behavior did not directly represent the output, are investigated in Section~\ref{sec:MNIST_readout} for the sake of comparison.
The datasets for training and those for testing to evaluate the performance were kept separate so that the generalization ability would be confirmable.
On the contrary, for the experimental results under the condition of using 100 specific samples for both training and testing, please refer to Section~\ref{sec:specific_MNIST}.
The purpose of this case was to check whether a clear division of roles between classification and control would emerge if the input was limited.
The training was conducted with a batch size of 50.

The procedures for creating label maps are as follows.
In the first place, when the movable mass points are placed in the proper positions to realize the best trajectory (Fig.~\ref{fig:MNIST} (b)), one noncentral mass point is selected and shifted in the $x$- and $y$-directions at distances of $\Delta x$ and $\Delta y$ from the initial position.
The other mass points are kept in the initial trained positions.
From this state, the transient dynamics of the MSDN is observed, and the actual and target trajectories of 10 types from zero to nine are compared.
The set of $\Delta x$ and $\Delta y$ is then classified as the label with the minimum error.
Regarding the mass point, the aforementioned procedure is repeated while changing $\Delta x$ and $\Delta y$ by a little so that the displacement covers a certain area.
The location of each displacement on the label map is finally color-coded according to its respective minimum error label.
Since there are 10 types of trained initial positions according to the 10 input labels, 10 label maps can be obtained for one mass point (Section~\ref{sec:label_maps}).

\subsection{Time-series emulation task}
In this task, the system takes an input and emulates the output of the target model based on the $y$-coordinate position of the CMP.
The input $u(t)$ was sampled from a uniformly independent and identical distribution $U(0,1)$, and scaled to $(0,20)$.
Unlike normal physical reservoir computing with readout layers, the output was embedded as a physical component.
Therefore, the roles of the readout layers were internalized into the body.
Section~\ref{sec:target_IPC} shows the IPC profiles of the target models used for the experiments and NARMA models.
For the body, the input is transformed at the brain and reflects the external forces on the movable mass points.
However, changing the external forces on the mass points at every simulation step (0.01 second) makes the time-series reconstruction difficult because each input value shifts to the next before its effect propagates through the MSDN.
Therefore, we introduced a variable $\tau$ so that the input would change at every $\tau$ simulation step~\cite{nakajima2013computing}; hence, the same input value was applied to the system, and the same external force acted on the mass points during $\tau$ steps.
The identical input continued to be entered for $\tau$ simulation time steps, and the states immediately after the final step were extracted as the output, when it was considered that the input had sufficiently propagated into the whole system.
Note that the physical simulation step size and the task time step size were different, as the width of the task time step was $\tau$ times larger.
In this experiment, $\tau$ was 20.
As a result, the input value switched every 0.2 seconds.

IPC is an index for measuring the information processing capability of a dynamical system.
From the perspectives of memory and nonlinearity, the input information retained in the states of the system can be examined.
Assuming that the current states are determined by the past input history, IPC evaluates the system by decomposing them into orthogonal polynomials consisting of past inputs.
The input $u(t)$ was converted to $(-1,1)$ by shifting and scaling and used.
The $n$th-order orthogonal polynomial is given as $\mathcal{F}_n\left(u(t-\Delta t)\right)$, where $\Delta t$ is a delay.
A polynomial with $n\geq2$ is equivalent to a nonlinear profile.
However, as mentioned above, the current system output value could also reflect the input at that time, and this was consistent with the IPC profile having the current input $u(t)$.
We can calculate the IPC even if the state to be decomposed is multidimensional, and the total capacity is equal to the rank.
Because the system output of this task was expressed as the $y$-axis position of the CMP, the output IPC resulted from the decomposition of the one-dimensional value into the polynomials of the input history. 
The IPC of each mass point was similarly calculated using each $x$-coordinate position or $y$-coordinate position as the state.

\subsection{Lissajous curve drawing task}
\label{sec:Lissajous_method}
Lissajous curves are defined as limit cycles that combine trigonometric functions. 
\begin{eqnarray}
  x&=&X\cos(\alpha t), \\
  y&=&Y\sin(\beta t+\delta).
\end{eqnarray}
The shapes of the curves are governed by $\alpha$, $\beta$, and $\delta$.
In this experiment, the variables were set to $(\alpha, \beta, \delta)=(1, 2, 0)$, and $X$ and $Y$ were set to $1/20$ of the diameter of the circular MSDN.

Noises added to the FL during training were sampled from a normal distribution with a mean of 0 and a standard deviation of 0.08.

Success in returning to the desired behavior was defined as
\begin{align}
  \frac{E}{E_{\text{orig}}}<2,
\end{align}
where $E$ was the error between the target and actual trajectories in the closed loop and $E_{\text{orig}}$ was the error in the open loop.

\subsection{Locomotion task}
\label{sec:locomotion_method}
A simple physical simulation that included external forces from the environment was prepared.
In this environment, gravity $\bm{F}_i^\text{grv}$, air resistance $\bm{F}_i^\text{air}$, and the reaction force from the ground $\bm{F}_i^\text{grd}$ act on the $i$th mass point.
They were implemented following a previous study~\cite{urbain2017morphological}. 
An $x$-$y$ coordinate system was assumed, and the $y$-direction was vertical to the ground.
\begin{align}
  \bm{F}_i^\text{grv}&=-m_i\bm{g}, \\
  \bm{F}_i^\text{air}&=-a\bm{v}_i, \\
  \bm{F}_i^\text{grd}&=[0,~c\exp(-cr_{iy})]^T,
\end{align}
where the gravitational acceleration $\bm{g}=[0,-9.81]^T$, proportionality factor of air resistance $a=0.1$, and ground reaction force constant $c=10$.
Note that the ground reaction force was formulated in this way to make the simulation differentiable, and it was possible that an object could sink into the ground.
In this simulation, the states of the body naturally change by interacting within the environment.
Therefore, the application of BPTSB, which is backpropagation through physics-based simulation, implies propagation that includes everything: the brain, body and environment.

The robot was supposed to move in the $x$-direction, and the mean squared error between the target and actual speeds was used for training.
The actual locomotion speed was the average of those of all mass points.
The target speed increased in accordance with the training time~\cite{hermans2014automated}.
\begin{align}
  v_\text{target}(t)=0.004t.
\end{align}
The parameters were updated every 200 time steps.

The noises used to train the FL were sampled from a normal distribution with a mean of 0 and a standard deviation of 0.2.

A normal distribution (with a mean of 0) provided the perturbation.
The perturbation was parameterized based on the standard deviation of the normal distribution, and the stability of the system was analyzed when the perturbation was in the range $[0.0, 1.0]$ in intervals of 0.1.
For each standard deviation, the noise was sampled as a perturbation using 100 different random seeds.
A successful return was defined as achieving a speed 0.7 times or more greater after the perturbation than the speed before the perturbation.

\section{Supplementary Information}
\subsection{Nonlinearity in linear mass-spring-damper networks}
\label{sec:MSDN_nonlinearity}
\begin{figure}[t]
  \centering
  \includegraphics[width=7cm,clip]{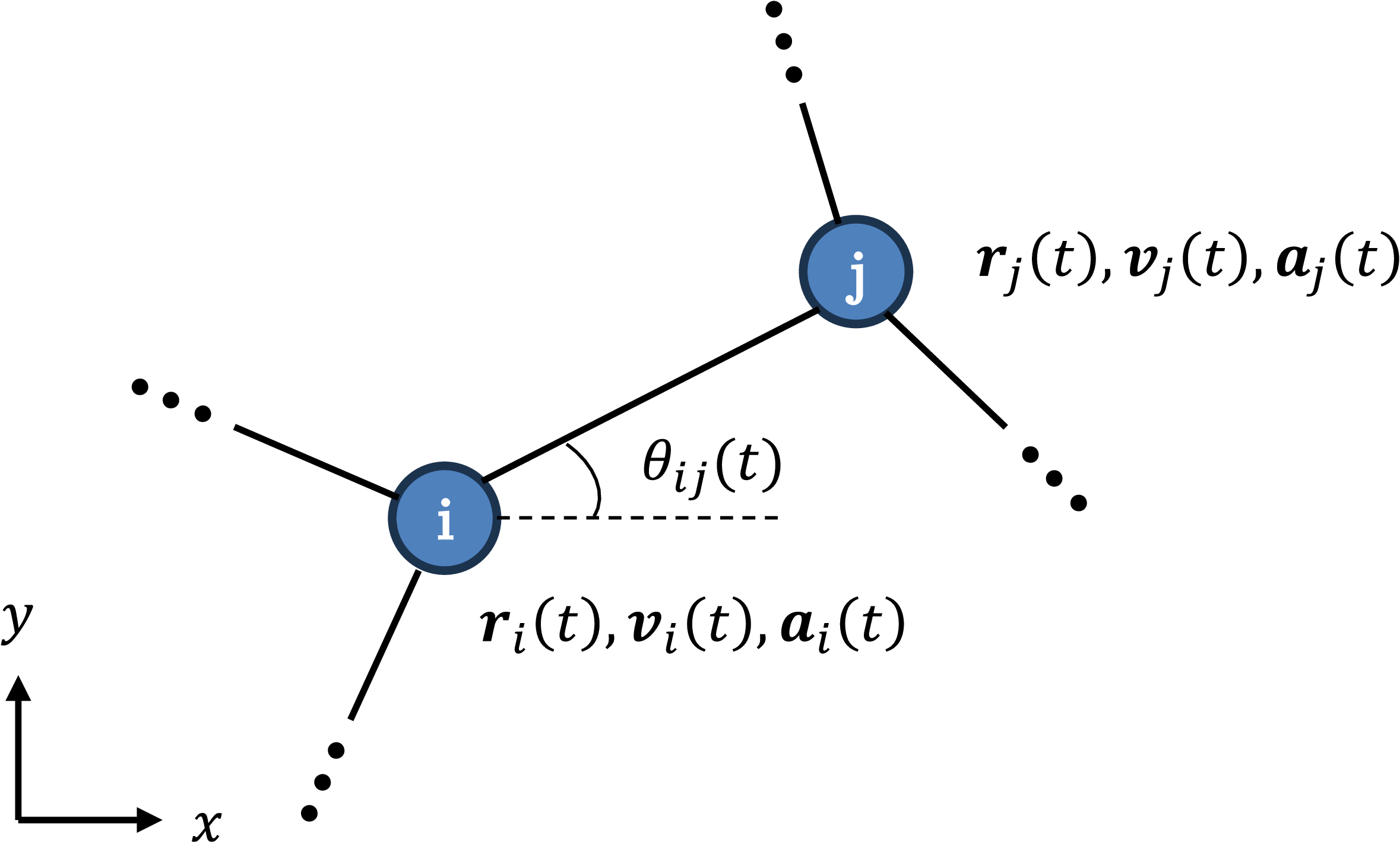}
  \caption{The arrangement of the mass points.}
  \label{fig:supp_mass-points_arrangement}
\end{figure}

In this study, although linear springs were used for body modeling, the mass-spring-damper networks (MSDNs) possessed nonlinearity because of a geometric factor described below.
This indicates that a system with a linear MSDN and a linear input layer (LIL) can generate nonlinear information processing capacity (IPC) profiles in a time-series emulation task.
In this section, focusing on a mass point $i$ (Fig.~\ref{fig:supp_mass-points_arrangement}), we prove the nonlinearity of the system.
The position, velocity, and acceleration of the mass point $i$ at time $t$ are described as $\bm{r}_{i}(t)$, $\bm{v}_{i}(t)$, and $\bm{a}_{i}(t)$, respectively.
The angle between the $i$th mass point and the $j$th mass point is expressed as $\theta_{ij}(t)$.
Let us assume that a one-dimensional input $u(t)$ is given at time $t$, and that the external force vector to which the input is transformed in the brain acts on each mass point.
The purpose of this section is to certify that MSDN can possess nonlinearity due to a geometric property in the spring connection, so we assume that the transformation in a brain is linear, i.e., the brain is defined as an LIL.
Therefore, $b_{ix}u(t)$ represents the $x$-direction component of an external force vector acting on the $i$th mass point, where $b_{ix}$ is one of the scalar values of the weights of the LIL. 

The Velocity Verlet algorithm~\cite{verlet1967computer} is described as follows:
\begin{align}
  \bm{r}(t+\Delta t)=\bm{r}(t)+\bm{v}(t)\Delta t+\frac{\bm{a}(t)}{2}(\Delta t)^2,\\
  \bm{v}(t+\Delta t)=\bm{v}(t)+\frac{\bm{a}(t+\Delta t)+\bm{a}(t)}{2}\Delta t.
\end{align}
The acceleration $\bm{a}(t)$ is calculated from an equation of motion.
Here, all masses $m$ were fixed at one for simplicity.
The relation between $\bm{a}(t)$ and force $F(t)$ is rewritten as
\begin{align}
  \bm{a}(t)=\frac{\bm{F}(t)}{m}=\bm{F}(t).
\end{align}
The resultant force of the spring forces and external forces, whose $x$-direction component is
\begin{align}
  F_{ix}(t)=-\sum_j \Bigl\{k_{ij}\left(r_{ix}(t)-r_{jx}(t)-l_{ij}\cos\theta_{ij}(t)\right)+d_{ij}\left(v_{ix}(t-\Delta t)-v_{jx}(t-\Delta t)\right)\Bigr\}+b_{ix}u(t),
\end{align}
acts on the $i$th mass point, resulting in the acceleration $a_{ix}(t)$:
\begin{align}
  a_{ix}(t)=-\sum_j \Bigl\{k_{ij}\left(r_{ix}(t)-r_{jx}(t)-l_{ij}\cos\theta_{ij}(t)\right)+d_{ij}\left(v_{ix}(t-\Delta t)-v_{jx}(t-\Delta t)\right)\Bigr\}+b_{ix}u(t),
\end{align}
where $k_{ij}$, $d_{ij}$, and $l_{ij}$ are the spring constant, damping coefficient, and rest length of the spring between the $i$th and the $j$th mass points, respectively.

We now prove that the physical simulation introduces nonlinear terms of input $u(t)$.
First, we consider the $x$-direction component of the position, velocity, and acceleration of the $i$th mass point at time $t+\Delta t$.
When the terms without $u(t)$ are represented as $\alpha_{ix}(t)$, the acceleration is
\begin{align}
  a_{ix}(t)=\alpha_{ix}(t)+b_{ix}u(t),
\end{align}
where
\begin{align}
  \alpha_{ix}(t)\coloneqq -\sum_j \Bigl\{k_{ij}\left(r_{ix}(t)-r_{jx}(t)-l_{ij}\cos\theta_{ij}(t)\right)+d_{ij}\left(v_{ix}(t-\Delta t)-v_{jx}(t-\Delta t)\right)\Bigr\}.
\end{align}
Hereafter, functions denoted by lower-case Greek letters are assumed to represent functions that summarize terms without inputs after time $t$.
Calculated based on the Velocity Verlet algorithm, the $x$ position is 
\begin{align}
  r_{ix}(t+\Delta t)&=r_{ix}(t)+v_{ix}(t)\Delta t+\frac{a_{ix}(t)}{2}(\Delta t)^2 \nonumber \\
  &=r_{ix}(t)+v_{ix}(t)\Delta t+\frac{(\Delta t)^2}{2}\alpha_{ix}(t)+\frac{(\Delta t)^2}{2}b_{ix}u(t) \nonumber \\
  &=\beta_{ix}(t)+\frac{(\Delta t)^2}{2}b_{ix}u(t),\\
  \beta_{ix}(t)&\coloneqq r_{ix}(t)+v_{ix}(t)\Delta t+\frac{(\Delta t)^2}{2}\alpha_{ix}(t).
\end{align}
The $x$ component of the acceleration at time $t+\Delta t$ is computed as
\begin{align}
  a_{ix}(t+\Delta t)={}&-\sum_j \Bigl\{k_{ij}\left(r_{ix}(t+\Delta t)-r_{jx}(t+\Delta t)-l_{ij}\cos\theta_{ij}(t+\Delta t)\right)+d_{ij}\left(v_{ix}(t)-v_{jx}(t)\right)\Bigr\}+b_{ix}u(t+\Delta t) \nonumber \\
  ={}&\gamma_{ix}(t)-\frac{(\Delta t)^2}{2}\sum_j k_{ij}(b_{ix}-b_{jx})u(t)-\sum_j l_{ij}\cos\theta_{ij}(t+\Delta t)+b_{ix}u(t+\Delta t),\\
  \gamma_{ix}(t)\coloneqq {}&-\sum_j \Bigl\{k_{ij}\left(\beta_{ix}(t)-\beta_{jx}(t)\right)+d_{ij}\left(v_{ix}(t)-v_{jx}(t)\right)\Bigr\}.
\end{align}
Here, the term $\cos\theta_{ij}(t+\Delta t)$ is written as 
\begin{align}
  \cos\theta_{ij}(t+\Delta t)&=-\frac{r_{ix}(t+\Delta t)-r_{jx}(t+\Delta t)}{\sqrt{\left(r_{ix}(t+\Delta t)-r_{jx}(t+\Delta t)\right)^2+\left(r_{iy}(t+\Delta t)-r_{jy}(t+\Delta t)\right)^2}} \nonumber \\
  &=-\frac{\beta_{ix}(t)-\beta_{jx}(t)+\frac{(\Delta t)^2}{2}(b_{ix}-b_{jx})u(t)}{\sqrt{\Bigl\{\beta_{ix}(t)-\beta_{jx}(t)+\frac{(\Delta t)^2}{2}(b_{ix}-b_{jx})u(t)\Bigr\}^2+\Bigl\{\beta_{iy}(t)-\beta_{jy}(t)+\frac{(\Delta t)^2}{2}(b_{iy}-b_{jy})u(t)\Bigr\}^2}}, 
\end{align}
which is a nonlinear term of the input $u(t)$.
Hence, we express this term as
\begin{align}
  \mathcal{N}\left(u(t)\right)\coloneqq \cos\theta_{ij}(t+\Delta t).
\end{align}
The acceleration at time $t+\Delta t$ is again described as follows:
\begin{align}
  a_{ix}(t+\Delta t)=\gamma_{ix}(t)-\frac{(\Delta t)^2}{2}\sum_j k_{ij}(b_{ix}-b_{jx})u(t)-\sum_j l_{ij}\mathcal{N}\left(u(t)\right)+b_{ix}u(t+\Delta t).
\end{align}
The $x$-direction component of the velocity at time $t+\Delta t$ is 
\begin{align}
  v_{ix}(t+\Delta t)={}&v_{ix}(t)+\frac{a_{ix}(t+\Delta t)+a_{ix}(t)}{2}\Delta t \nonumber \\
  ={}&v_{ix}(t)+\frac{\Delta t}{2}\Bigl\{\alpha_{ix}(t)+b_{ix}u(t)+\gamma_{ix}(t)-\frac{(\Delta t)^2}{2}\sum_j k_{ij}(b_{ix}-b_{jx})u(t)-\sum_j l_{ij}\mathcal{N}\left(u(t)\right)+b_{ix}u(t+\Delta t)\Bigr\} \nonumber \\
  ={}&\delta_{ix}(t)+\frac{\Delta t}{2}\Biggl\{b_{ix}-\frac{(\Delta t)^2}{2}\sum_j k_{ij}(b_{ix}-b_{jx})\Biggr\}u(t)-\frac{\Delta t}{2}\sum_j l_{ij}\mathcal{N}\left(u(t)\right)+\frac{\Delta t}{2}b_{ix}u(t+\Delta t),\\
  \delta_{ix}(t)\coloneqq {}&v_{ix}(t)+\frac{\Delta t}{2}\left(\alpha_{ix}(t)+\gamma_{ix}(t)\right),
\end{align}
and it has a term that includes $\mathcal{N}\left(u(t)\right)$.
Then, considering the position at the next step, we can calculate it as 
\begin{align}
  r_{ix}(t+2\Delta t)={}&r_{ix}(t+\Delta t)+v_{ix}(t+\Delta t)\Delta t+\frac{a_{ix}(t+\Delta t)}{2}(\Delta t)^2 \nonumber \\
  ={}&\epsilon_{ix}(t)+\Biggl\{(\Delta t)^2b_{ix}-\frac{(\Delta t)^2}{2}\sum_j k_{ij}(b_{ix}-b_{jx})\Biggr\}u(t)-(\Delta t)^2\sum_j l_{ij}\mathcal{N}\left(u(t)\right)+(\Delta t)^2b_{ix}u(t+\Delta t),\\
  \epsilon_{ix}(t)\coloneqq {}&\beta_{ix}(t)+\Delta t\delta_{ix}(t)+\frac{(\Delta t)^2}{2}\gamma_{ix}(t),
\end{align}
which shows that the position of the mass point can have the nonlinear terms of the past input $u(t)$.
The same is true for the $y$-direction.

It was shown that even systems that use linear springs can possess nonlinearity depending on the geometric characteristics of the network structure.
Note that the nonlinear factors in this system are trigonometric terms caused by the two-dimensional arrangement of connected mass points.
Therefore, if mass points are located in one dimension and move one-dimensionally, for example, if they connect in series and move only in such direction, the system remains linear.

\subsection{MNIST classification task}

\subsubsection{MNIST classification by reading states out}
\label{sec:MNIST_readout}
While the experiment in Section~\ref{sec:MNIST} worked with MNIST classification by regarding physical behaviors as the output, we also tackled this task using a standard method of physical reservoir computing.
The classification was performed by reading out the states of the system with linear weights.
The aim of this experiment was to investigate the effect on task performance of incorporating the physical dynamics of MSDNs into a part of the training, compared with the case in which only a feed-forward neural network (FNN) performed the classification.
We adopted an LIL as a brain, and an MSDN with double-circle structures ($N_\text{mov}=17,65$) or multiple-circle structures ($N_\text{mov}=17,65$) as a body.
We also prepared a linear readout layer on the output side, which can be considered a part of the brain.
The brain converted the input into the initial positions of the movable mass points, and then the body produced transient dynamics based on the initial arrangement without any external forces.
The transient period was limited to 1 second, and the states of the body were stored every 0.1 seconds.
The positions of the movable mass points ($2N_\text{mov}$ dimensions) were designated as the states.
After a one-second transient time, the classification was performed by converting a $20N_\text{mov}$-dimensional vector composed of the stored states into the output using the linear readout layer.
A softmax function and cross entropy loss were used for the training.
In addition, we trained a system without bodies, which classified inputs by reading out the output directly from the input layer.
This system is equivalent to a two-layer linear neural network.

\begin{table}[t]
    \centering
    \caption{Classification accuracy obtained by reading states out.}
    \begin{tabular}{c|c|c|c|c|c}
        \hline
        & ~Only brain~ & \multicolumn{2}{c|}{~Double-circle~} & \multicolumn{2}{|c}{~Multiple-circle~} \\
        \hline
        $N_\text{mov}$ & - & 17 & 65 & 17 & 65 \\
        \hline
        Accuracy [\%] & 91.18 & 95.69 & 96.77 & 95.36 & 96.22 \\
        \hline
    \end{tabular}
    \label{tab:MNIST_readout}
\end{table}

Table~\ref{tab:MNIST_readout} shows the classification accuracy of the five types of systems: one without bodies (a linear neural network), double-circle structures ($N_\text{mov}=17,65$), and multiple-circle structures ($N_\text{mov}=17,65$).
These systems were optimized for 100 epochs.
The system without bodies was attributed 91.18\% accuracy.
On the other hand, the systems that included bodies achieved higher than 95\% accuracy, indicating the usefulness of incorporating physical dynamics into the classification.
As shown in Section~\ref{sec:MSDN_nonlinearity}, the system with a body is nonlinear.
Regardless of whether the structure of the body had two circles or multiple circles, there was a pattern in which the larger the number of system states, the higher the classification accuracy.
This is presumed to be because the larger system size generated abundant dynamics inside the system.
This experimental setting did not utilize the physical behavior directly, so the differences in body structure had little effect on classification performance when the number of system states was the same.

\subsubsection{MNIST classification with specific inputs}
\label{sec:specific_MNIST}
In Section~\ref{sec:MNIST}, classification performance was evaluated with a dataset that was not used for the training.
In the experiment reported here, we examined the classification task using only 100 input images (10 pieces for each number from zero to nine), which were randomly chosen from the MNIST training dataset.
These inputs were used for both training and testing; the training was performed under conditions that allowed overfitting regardless of generalization.
The purpose of this experiment was to confirm whether a clear division of the roles of classification and control between the brain and the body would emerge if the input was specifically defined.
Although the experiment in Section~\ref{sec:MNIST} did not involve a clear division of roles, it was assumed that this setup can easily bring this about because the system has to distribute each of 10 types of input to 10 types of output.
If there is a division of roles in which the brain completely classifies the input numbers and the body is responsible only for drawing, exactly the same information must be delivered from the brain to the body based on the identical input label.
In short, the initial positions of the mass points should match for each label, resulting in a subsequent trajectory that is limited to 10 types.
An LIL and a multilayer perceptron (MLP) were used for this experiment as a brain, and the body had four structural varieties, including double-circle structures ($N_\text{mov}=17,65$) and multiple-circle structures ($N_\text{mov}=17,65$).

\begin{figure*}[t]
    \centering
    \includegraphics[keepaspectratio,width=\columnwidth,clip]{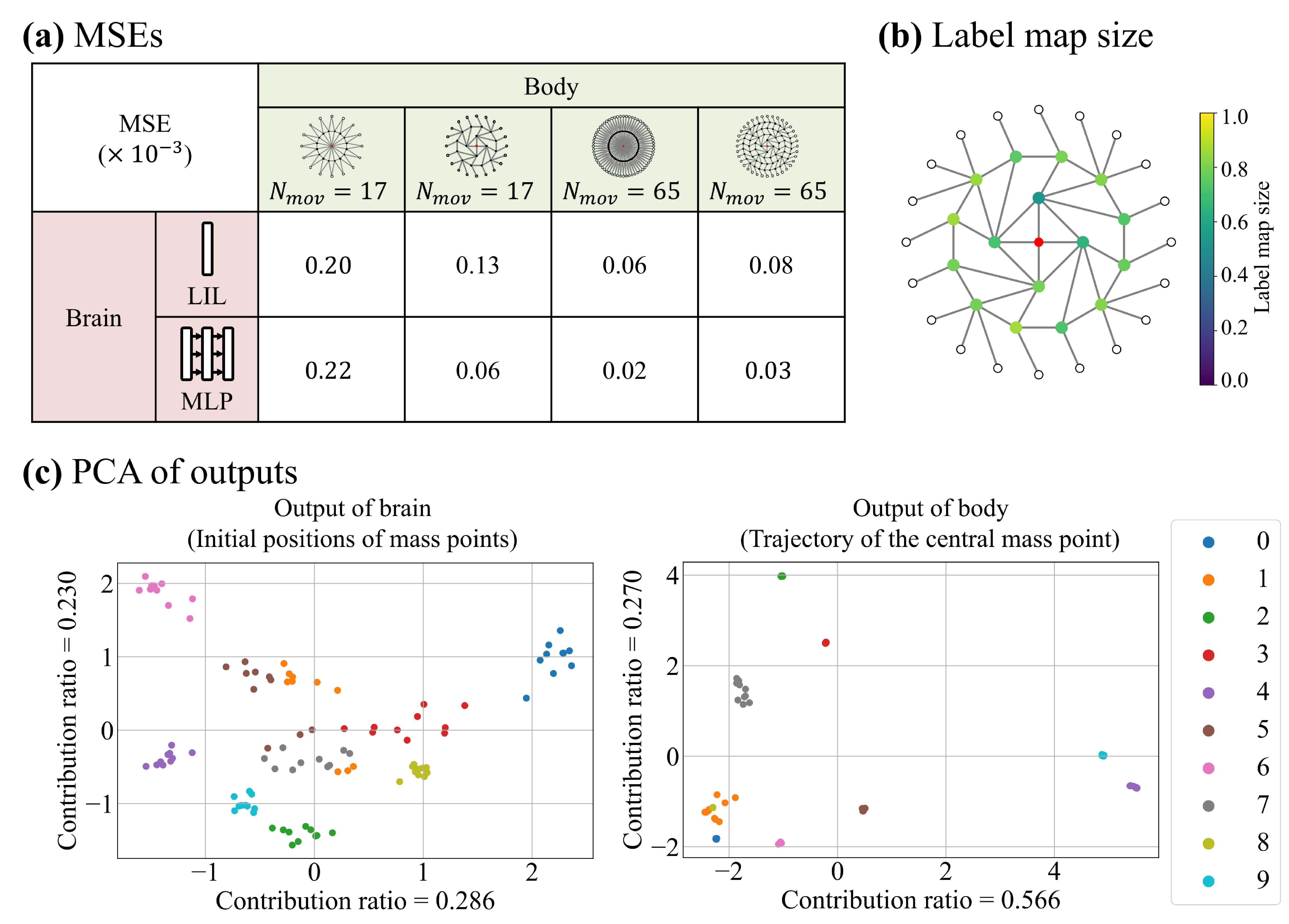}
    \caption{MNIST classification with specific inputs.
    (a) Averaged mean squared errors of the drawn trajectories.
    The 100 input MNIST images are used for both training and testing.
    (b) Degrees of simplicity of the label map structures.
    The smaller the value, the more sensitive the mass point to a change in initial position.
    (c) The outputs of the brain (the initial positions of the movable mass points) and the body (the trajectory of the CMP) are compressed into two dimensions using PCA.
    Each color corresponds to an input label.
    The greater the differences in color, the stronger the label classification.}
    \label{fig:supp_MNIST}
\end{figure*}

With the specific 100 images used for training and testing, all systems achieved high accuracy in drawing the label numbers (Fig.~\ref{fig:supp_MNIST} (a)).
The MLP yielded less error than the LIL, and bodies with more mass points performed better.
The label map sizes of the multiple-circle structure ($N_\text{mov}=17$; Fig.~\ref{fig:supp_MNIST} (b)) indicate that the values of the mass points directly connected to the central mass point (CMP) are comparatively small, which was the same result reached in Section~\ref{sec:MNIST}.
Fig.~\ref{fig:supp_MNIST} (c) represents the results of principal component analysis (PCA) processing of the output of the brain and the body.
As for the result for the brain, there is moderate separation by labels.
This result indicates that the brain did not deliver a unique output to the body in accordance with the input label.
On the other hand, the result for the body demonstrates explicit separation for each label.
This suggests that not only the brain but also the body contributed to the classification of the MNIST images.
In other words, no clear division of roles between the brain and the body was produced even under the condition that the input images were confined, and the system was optimized complementarily.

\subsubsection{Label maps}
\label{sec:label_maps}

\begin{figure*}[t]
    \centering
    \includegraphics[keepaspectratio,width=\columnwidth,clip]{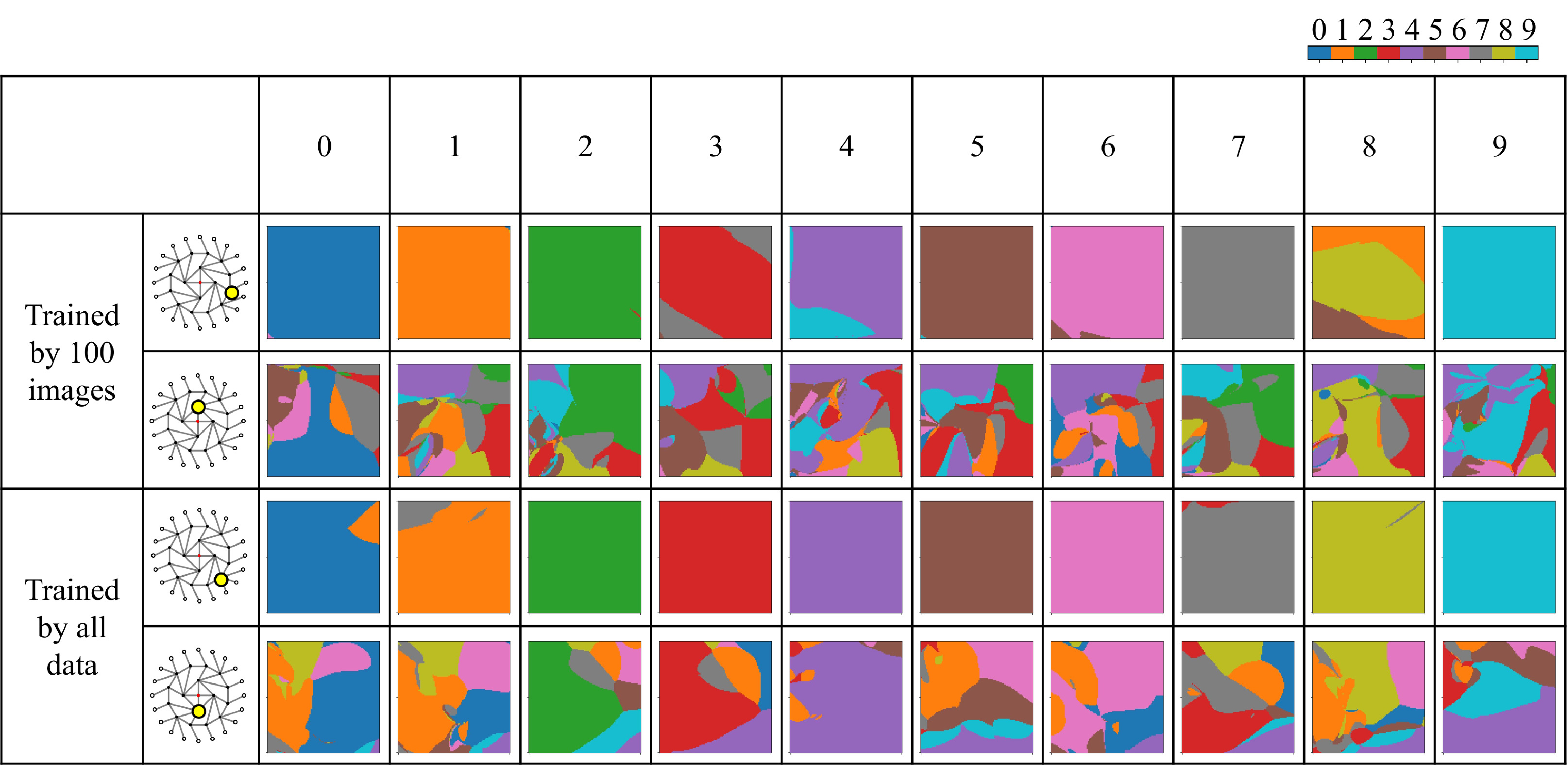}
    \caption{Label maps.
    Label maps with the highest label map size and the lowest one are displayed.
    The horizontal and vertical axes correspond to $\Delta x$ and $\Delta y$, respectively, which are equivalent to the distances the mass point of interest is shifted from its initial position.
    Top two rows: Maps of the system trained with the specific inputs.
    Bottom two rows: Maps of the systems trained with all data.
    Each color corresponds to the associated label.}
    \label{fig:supp_labelmap}
\end{figure*}

The label maps of the mass points with the maximum and minimum label map sizes for a system trained using only 100 specific images and a system trained using all the training data in Fig.~\ref{fig:supp_labelmap}.
These systems have an MLP and a multiple-circle-structure MSDN ($N_\text{mov}=17$).
The mass point with the maximum label map size does not connect to the CMP directly; on the contrary, the point with the minimum size has a direct connection.
The system trained with specific inputs has more complex and intricate map structures.
This is thought to be attributable to the limited pattern of inputs, which resulted in poor generalization performance and caused the sensitivity to small changes in the inputs.
On the contrary, this result can be explained by the body acquiring a generalization ability that allowed it to absorb some degree of the fluctuation in the brain's output by training with the entire dataset and thus suppressing a complication of label maps.

\subsection{Time-series emulation task}

\subsubsection{Memory task}
\label{sec:memory_task}
We verified the properties of a system's memory in this task.
A brain was set to be an FNN, so there was no mechanism for retaining memories in the brain itself.
Whether the brain is an LIL or an MLP has no effect on the investigation of memory.
Hence, the brain was fixed as an LIL, and it reconstructed the past input together with a body.
Either a double-circle structure ($N_\text{mov}=17$) and a multiple-circle structure ($N_\text{mov}=17$) was used as bodies.

When a variable $\Delta t$ represents a delayed time step, the target output to be emulated by the system output is the input value before the $\Delta t$ time step:
\begin{align}
  y(t)=u(t-\Delta t).
\end{align}
The system has to reproduce the past input as is, so no nonlinearity is required.
Note that the term ``time step" here indicates not the physical simulation time step but the task time step, in which the input value switches once.

\begin{figure*}[t]
  \centering
  \includegraphics[width=\linewidth,clip]{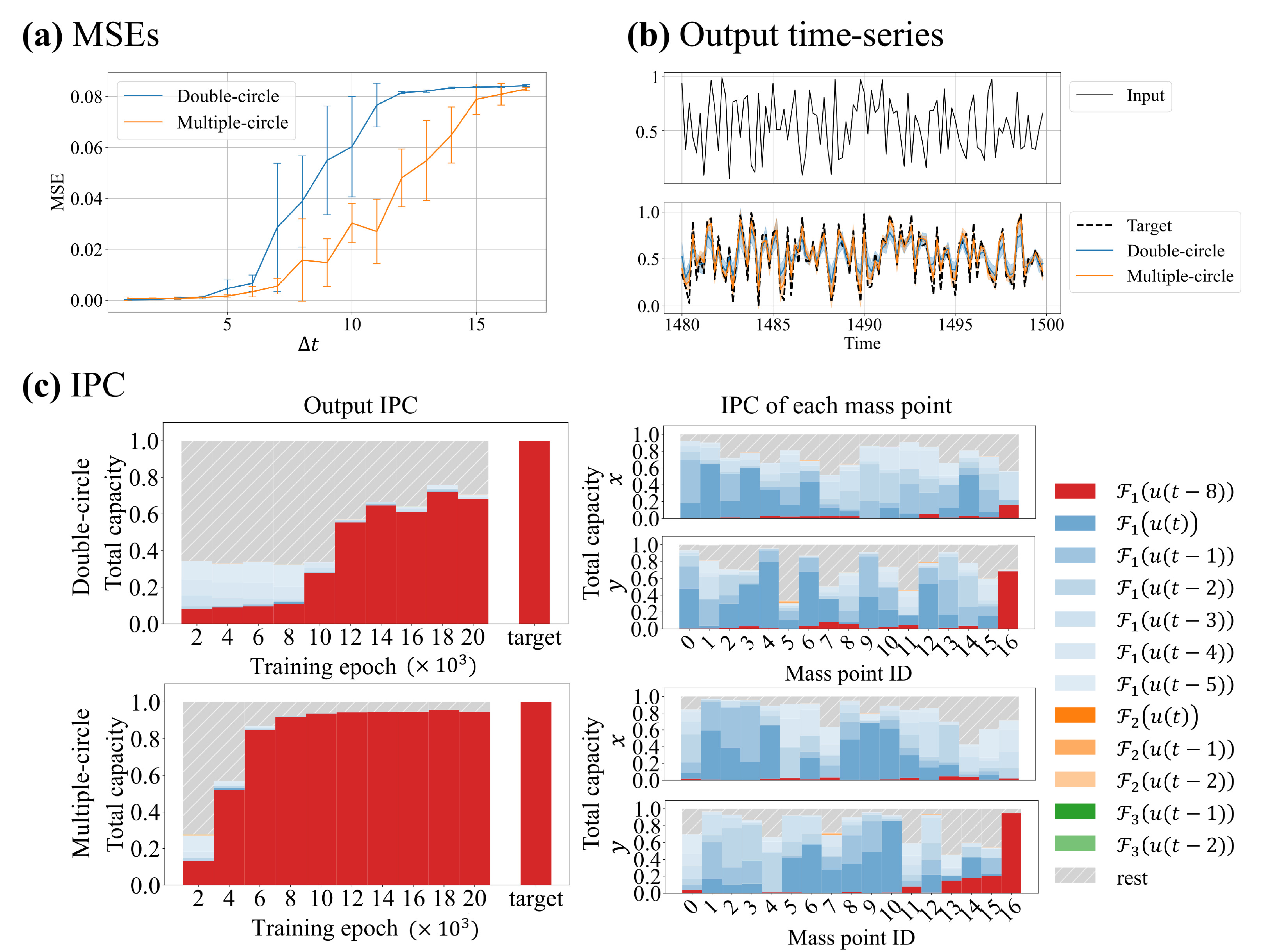}
  \caption{The memory task.
  (a) Reconstruction error value when $1\leq \Delta t\leq17$.
  The error bars correspond to the standard deviations.
  (b) The input time-series is shown in the top figure, while the target and output time-series are shown in the bottom figure ($\Delta t=8$).
  (c)  The left side represents the IPC values of the system output, which are arranged from left to right according to training epochs.
  The rightmost bar represents the IPC value of the target time-series.
  On the other hand, the right figures represent the IPC values of each movable mass point (one dimension each in the $x$ and $y$ axes; $\Delta t=8$).
  The degree of the orthogonal polynomial and the delayed time step of the input are given as $n$ and $\Delta t$, respectively, in $\mathcal{F}_n\left(u(t-\Delta t)\right)$, and they are equivalent to the color and the intensity of its bars.
  The color of the IPC profile $\mathcal{F}_1\left(u(t-8)\right)$, which occupies all of the IPC of the target output, is illustrated in red for clarity.
  The total capacity reached the rank of one in all cases.}
  \label{fig:supp_memory-task}
\end{figure*}

We conducted the memory task with the delayed time step $1\leq \Delta t\leq17$ (Fig.~\ref{fig:supp_memory-task} (a)).
The larger the delayed time step $\Delta t$, the larger the reconstruction error, regardless of the body shape.
The systems with a multiple-circle structure generally tended to have smaller errors than those with a double-circle structure, suggesting superiority in terms of memory properties.
Fig.~\ref{fig:supp_memory-task} (b) shows the time-series of the input, the target output, and the system output when $\Delta t=8$.
The system with multiple circles can emulate the target time-series more accurately.
Fig.~\ref{fig:supp_memory-task} (c) displays the information processing capacity (IPC)~\cite{dambre2012information} of the output and the positions of each movable mass point.
In the double-circle structure, the output IPC does not equal the target IPC even after sufficient training epochs, while the multiple-circle body can reproduce most of it.
Focusing on the IPC of each mass point in the multiple-circle body, we can see that the profiles of the time delay $\Delta t=8$ at the mass points with IDs 14, 15, and 16 are relatively large.
These mass points are located at the innermost part of the circles and connected to the CMP directly, thus suggesting that geometric features of the body affect the IPC profiles. 

\subsubsection{The target IPC}
\label{sec:target_IPC}

\begin{figure*}[t]
    \centering
    \includegraphics[keepaspectratio,width=0.9\columnwidth,clip]{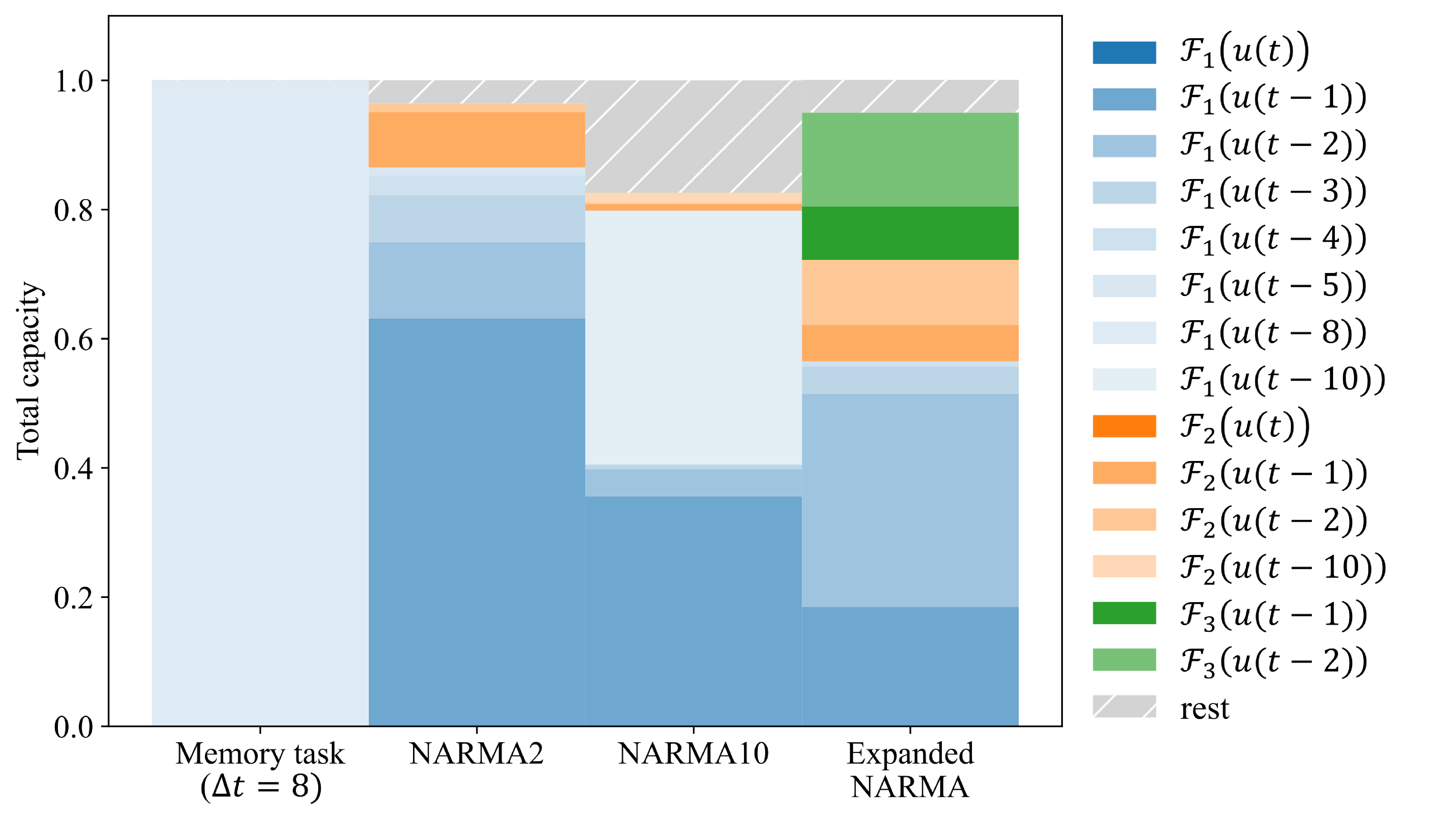}
    \caption{The target IPC.}
    \label{fig:supp_target_ipc}
\end{figure*}

Nonlinear autoregressive moving average (NARMA)~\cite{atiya2000new} is a benchmark task at the measure memory properties and nonlinearity in recurrent neural networks.
The NARMA model is calculated as follows:
\begin{align}
  y(t+1)=
  \begin{cases}
    ~0.4y(t)+0.4y(t)y(t-1)+0.6u^3(t)+0.1 & (m=2),\\
    \displaystyle ~0.3y(t)+0.05y(t)\sum_{i=0}^{m-1}y(t-i)+1.5u(t-m+1)u(t)+0.1 & (m>2),
  \end{cases}
\end{align}
where $u(t)$ and $y(t)$ represent the input and the state at time $t$, and the output time-series is formed by the time-series of $y$.
In particular, NARMA2 ($m=2$) and NARMA10 ($m=10$) models have been adopted in many studies.
However, it has been reported that the ratio of higher-order nonlinearity in NARMA models is small, and a simple linear system can frequently exhibit good performance~\cite{kubota2021unifying}.
In Section~\ref{sec:timeseries}, for the purpose of enhancing higher-order nonlinearity, we defined the target model as follows:
\begin{align}
  y(t+1)=0.3y(t)+\sum_{i=0}^{m-1}\left(u(t-i)^3-u(t-i)^4\right)+0.2\sum_{i=0}^{m-1}\left(y(t-i)^3-y(t-i)^4\right)+0.1.
\end{align}
We call this model the expanded NARMA model.
The IPC values of two types of NARMA models (NARMA2 and 10), the expanded NARMA model, and the target model of the memory task ($\Delta t=8$) are displayed in Fig.~\ref{fig:supp_target_ipc}.
In the target for the memory task, the IPC consists only of the profile in focus, which indicates that the output can be reproduced only by the delayed input.
In the NARMA models, first-order and one-step delayed profiles are dominant.
The 10-step delayed profile also stands out in NARMA10.
However, the profiles with second-or-more-order nonlinearity have small rations.
On the other hand, the proportions of second- and third-order nonlinear profiles are relatively larger in the expanded NARMA model, suggesting the existence of sufficient nonlinearity.
Therefore, Section~\ref{sec:MNIST} was able to focus on the higher-order nonlinearity rather than the original NARMA task.

\subsection{Tasks with sine wave generator}
In this section, we introduce several experiments with training systems composed of a sine wave generator (SWG) and an MSDN to acquire a single behavior or switch behaviors.

\subsubsection{Behavioral switching}
\label{sec:switching}
\begin{figure*}[t]
  \centering
  \includegraphics[keepaspectratio,width=\linewidth,clip]{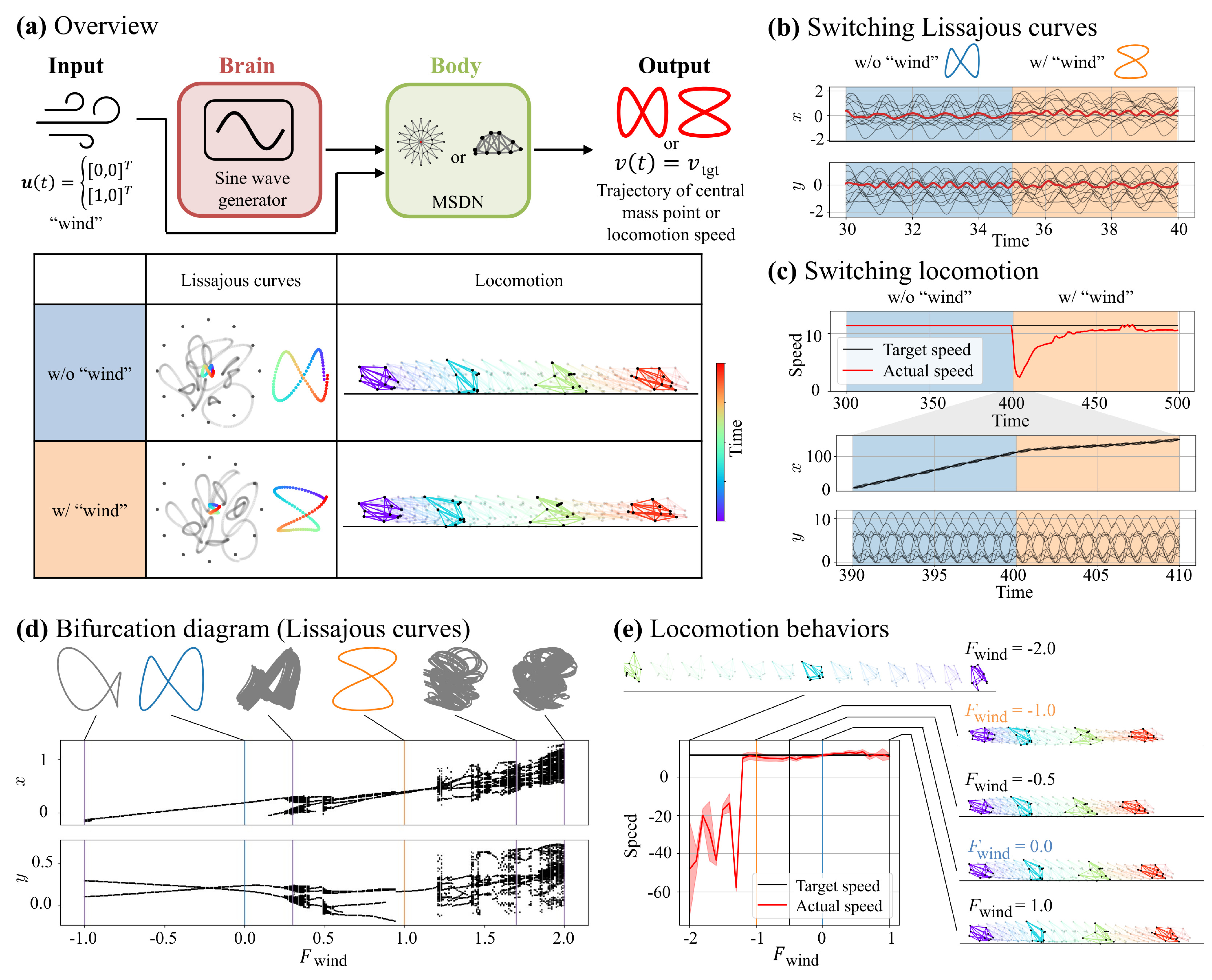}
  \caption{Training behaviors of an SWG.
    (a) The input is the external force of \wind in the $x$-direction, and the output is the trajectory of the CMP or locomotion speed.
    In the Lissajous curve drawing task, the training was arranged that the trajectory would become a Lissajous curve.
    The drawn curve was switched under the influence of \wind (Supplementary Video 5).
    On the other hand, in the locomotion task, a robot was optimized to move forward at the target speed regardless of $F_{\text{wind}}$ (Supplementary Video 6).
    The bottom table shows the resultant behaviors, where the color corresponds to elapsed time.
    (b) The dynamics of each movable mass point (the red line represents that of the CMP) before and after the switch.
    $F_{\text{wind}}=0$ is represented as blue and $F_{\text{wind}}=1$ as orange.
    (c) The top graph represents the target and actual speed, and the bottom graph shows each mass point's dynamics.
    (d) In the system that learned to switch Lissajous curves according to $F_{\text{wind}}$, the respective maxima in the $x$- and $y$-directions of the CMP are plotted when $F_{\text{wind}}$ varies between $[-1.0,2.0]$ (a bifurcation diagram).
    The trajectory drawn at each $F_{\text{wind}}$ value are shown at the top of the bifurcation diagram.
    (e) In the system that learned forward locomotion, the locomotion speed is plotted when $F_{\text{wind}}$ varied between $[-1.0,2.0]$.
    In addition, the locomotion behaviors are depicted as color-coded over time when $F_{\text{wind}}$ is -2.0, -1.0 (trained), -0.5, 0.0 (trained), and 1.0 (Supplementary Video 6).}
    \label{fig:supp_sine-wave-generator}
\end{figure*}

First, we took on a Lissajous curve drawing task.
A system composed of a double-circle MSDN ($N_\text{mov}=13$) was trained such that the CMP drew a Lissajous curve.
In other words, the objective was to embed a limit cycle into a brain--body coupling system.
This experiment also involved behavior switching according to an external input by embedding plural curves into the system (Fig.~\ref{fig:supp_sine-wave-generator} (a)).
For a physical-system-specific conditioning, we defined blowing \wind as the external inputs.
In particular, blowing \wind was equal to a constant external force being applied all mass points from a constant direction.
The direction of the blowing \wind was set as the $x$-axis, and $F_{\text{wind}}$ represented the magnitude of the \windper
During the optimization, $F_{\text{wind}}$ was dealt with as binary.
The times at which the \wind started and stopped were set randomly throughout the training.
By switching the target trajectories in accordance with the input changes, the system was updated to capture the ability to change what was being drawing.
The target trajectory was displaced in the $x$-direction when there was \windper
The error was calculated as the mean squared error between the actual trajectory of the CMP and the target.
What should be noted here is that we trained the system by having it search for the appropriate physical and control parameters of the trajectory of the CMP, instead of telling it the correct way to drive springs.
Lissajous curves are defined as limit cycles that combine trigonometric functions:
\begin{eqnarray}
  x&=&X\cos(\alpha t), \\
  y&=&Y\sin(\beta t+\delta).
\end{eqnarray}
The shapes of the curves are governed by $\alpha$, $\beta$, and $\delta$.
The variables $X, Y$ were set to $1/20$ of the diameter of the circular MSDN.

In the first place, the system was trained to draw a single Lissajous curve if $F_{\text{wind}}=0$.
By modulating all the springs of the body through signals from the brain, the Lissajous curve of $(\alpha,\beta,\delta)=(1,2,0)$ was successfully embedded into the brain--body coupling system.
This means that the CMP succeeded in continuing to draw the target Lissajous curve based on the sine wave modulation.

We then experimented with whether it could switch what it was drawing under the influence of the \windper
We prepared a system that was trained to draw a single curve in advance and optimized it for switching behaviors through additional training.
The additional curve was constituted by $(\alpha,\beta,\delta)=(2,1,\pi/2)$.
This curve is point symmetric with the curve represented by $(\alpha,\beta,\delta)=(1,2,0)$.
As a result, the embedding of the plural Lissajous curves into the system was successful, although the error between the drawn trajectory and the target grew larger after additional learning (Fig.~\ref{fig:supp_sine-wave-generator} (a)).
The \wind acted as a switching trigger reflecting the external forces on the respective mass points.
As shown in Fig.~\ref{fig:supp_sine-wave-generator} (b), the trajectory of the CMP achieved binary switching, and the dynamics also changed.
It could be confirmed that the CMP frequencies (the red lines) had been swapped between the $x$-direction and the $y$-direction by the \windper
Note that there was no periodical external signal or force, which means that each mass point was subject to a constant force in a constant direction if $F_{\text{wind}}=1$.
Moreover, after the parameter optimization, the trainable physical or control parameters were kept constant.
Thus, the switching behavior was realized by the brain, which was generating constant signals and the body, which had constant physical properties.

$F_{\text{wind}}$ was set to be a binary value, zero or one, during optimization.
In this regard, it has been shown that global bifurcation structures can be inferred based only on learning for limited local states ~\cite{kim2021teaching,flynn2021multifunctionality,kong2021machine}.
Our experiment could therefore provide valuable insight into the field from a physical-behavioral point of view.
Thus, we analyzed the system behaviors when the external input value for the system was unknown, verifying an interpolability and an extrapolability aquired while training.
We varied $F_{\text{wind}}$ in the range of $[-1,2]$ in increments of 0.05 and derived a bifurcation diagram from the behavior (Fig.~\ref{fig:supp_sine-wave-generator} (d)).
The blue and orange areas correspond to $F_{\text{wind}}$ values of zero and one.
Various bifurcation structures were observed: A stable trajectory was drawn from -1 to around 0.2, but chaotic behaviors were observed when $F_{\text{wind}}\in (0.2,0.6), (1.2,2.0)$.
An important conclusion derived from this analysis is that it is possible for there to be more diverse dynamics embedded in brain--body coupling systems than they have learned.
The embedded dynamics is caused by the interaction between the learning target and the system itself, so it may be possible to predict or control what happens by properly accounting for the nonlinear dynamics.
In other words, the use of such diverse bifurcation structures suggests an important contribution to the control of qualitatively different patterns ~\cite{akashi2024embedding,kabayama2024designing,terasaki2024thermodynamic}.

Using similar settings, we demonstrated the framework in locomotion in the context of robotics.
A simple caterpillar-shaped MSDN was used as a body.
The numbers of mass points and springs were fixed at 10 and 24.

First, the robot, which was composed of an SWG and an MSDN, was trained to move forward via backpropagation through soft body (BPTSB).
The robot was supposed to move in the $x$-direction, and the mean squared error between target and the actual speed was used for training.
The actual locomotion speed was the average of those of all the mass points.
Before optimization, an initialized system produced wriggling behaviors by modulated springs.
Optimization of multiple systems separately initialized with different random seeds led to quantitatively diverse locomotion styles.
In all systems, simultaneous training of the brain and the body was performed to optimize an agent moving forward on a plane.

In the locomotion task, behavioral switching was also attempted.
As in the Lissajous curve drawing task, locomotion was required to switch according to the \windper
In particular, the agent had to keep a constant locomotion speed whether a headwind was blowing or not.
When a headwind was blowing, a constant external force was exerted backwards on each mass point.
Therefore, it was necessary to take action to change the propulsion force according to $F_{\text{wind}}$ in order to maintain a constant speed, which corresponds to locomotion switching.
The training procedure was as follows:
\begin{enumerate}
  \item A brain--body coupling system is trained to move forward. The acquired maximum speed is $v_{\text{max}}$.
  \item The target speed is set to $v_{\text{tgt}}=v_{\text{max}}/2$. This value was arbitrarily determined as a feasible speed.
  \item The system parameters are updated by propagating error backwards. The error is calculated based on the target speed $v_{\text{tgt}}$ and the actual speed, and the \wind blows periodically.
\end{enumerate}
The third step was additional training.
As a result, we obtained an agent that maintained constant locomotion regardless of the absence or presence of \wind ($F_{\text{wind}}=0,-1$; Fig.~\ref{fig:supp_sine-wave-generator} (a)).
As shown in Fig.~\ref{fig:supp_sine-wave-generator} (c), the locomotion speed began to slow down immediately after the \wind started blowing, but it progressively increased and returned to locomotion equivalent to windless conditions.
Hence, this suggested that the locomotion was switched by \wind as a trigger.
Fig.~\ref{fig:supp_sine-wave-generator} (e) represents the target and actual speed when $F_{\text{wind}}\in [-2,1]$. 
With the \wind ($-1.0<F_{\text{wind}}<1.0$), which was unknown to it, the agent accomplished locomotion at a speed somewhat close to the target even through the $F_{\text{wind}}$ was changing.
This result means that interpolation was performed in an unlearned parameter region.
Qualitatively, it was assumed that brakes were applied using a mass point on the front of the body by changing the posture according to $F_{\text{wind}}$: The stronger the tailwind, the more the robot leaned forward.
On the other hand, the locomotion speed was far from the target speed when $F_{\text{wind}}<-1.0$.
This was because the robot floated in the air due to the \wind if the headwind was strong.

\subsubsection{Switching timing}
\label{sec:switching_timing}
During training, the trigger for switching was the time when the \wind started blowing and the time when it stopped blowing.
In the experiment above, they were set to be indeterminate, that is, the timing for switching the \wind and the period of the control signal from the SWG did not match.
The rest length of a spring is modulated as
\begin{align}
  l_{ij}^\text{M}(t)=l_{ij}\left(1+A_{ij}\sin(\omega t+\phi_{ij})\right).
\end{align}
If the system had the ability to draw a Lissajous curve, the CMP of the system passed through the same spot on the trajectory every time $\frac{2\pi}{\omega}$.
The switching timing was determined by dividing the period of the sine wave into 100 and randomly selecting one of these divisions.
On the basis of a given time $t_0$, the timing for starting blowing or stopping blowing $t_\text{wind}$ is described as follows:
\begin{align}
  t_\text{wind}=t_0+\frac{2\pi}{\omega}\frac{n}{100},
\end{align}
where $n\in [0,99]$ is a random integer.

\begin{figure*}[t]
  \centering
  \includegraphics[width=\linewidth,clip]{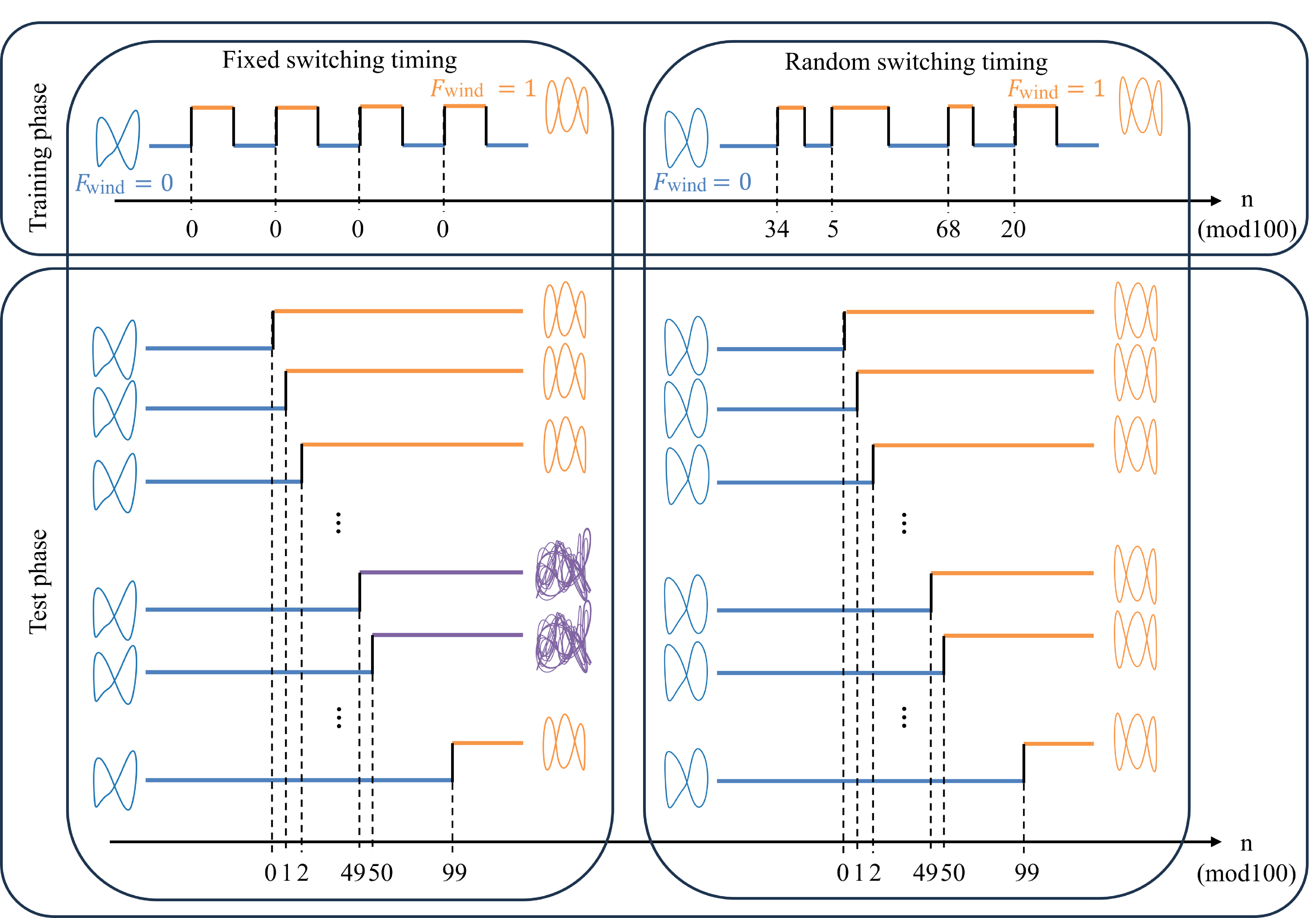}
  \caption{System behaviors with fixed or random timing for switching.
  The drawn trajectories represent when the timing for switching \wind is fixed (left) and random (right).
  Each line stands for the magnitude of the \windcom and the blue area corresponds to zero (without \windpar, while orange corresponds to one (with \windpar.
  Each actual trajectory is also attached, and the trajectories equivalent to the target are in blue or orange, while the others are designated in purple.}
  \label{fig:supp_lissajous_switching}
\end{figure*}

In addition to the system trained by presenting switching conditions at various times, we trained a system with a fixed switching period.
Using a format that performs additional training on a system that has trained for Lissajous curve ($(\alpha,\beta,\delta)=(1,2,0)$) drawing, we constructed a system to draw one ($(\alpha,\beta,\delta)=(1,2,0)$) or another ($(\alpha,\beta,\delta)=(1,3,0)$) according to \windper
A fixed value ($n=0$) was used to train the system with a fixed switching time.
As a result, the behavioral switching was achieved under training conditions for both systems.
We then confirmed the behaviors of the CMP by changing $n$ comprehensively (Fig.~\ref{fig:supp_lissajous_switching}), which was equivalent to the change in the timing of the switch.
The trajectory of the system trained with random switch timings could converge to the trained trajectory no matter when the \wind switched.
On the other hand, the trajectory of the system trained with a fixed switching timing converged to different attractors under conditions where \wind was blowing when $22\leq n\leq69$.
This result suggests not only the need to present diverse conditions during training when embedding a robust behavior, but also the existence of untrained attractors.

\subsubsection{Simultaneous training for a closed loop}
\label{sec:CL_BPTSB}

\begin{figure*}[t]
  \centering
  \includegraphics[keepaspectratio,width=\linewidth,clip]{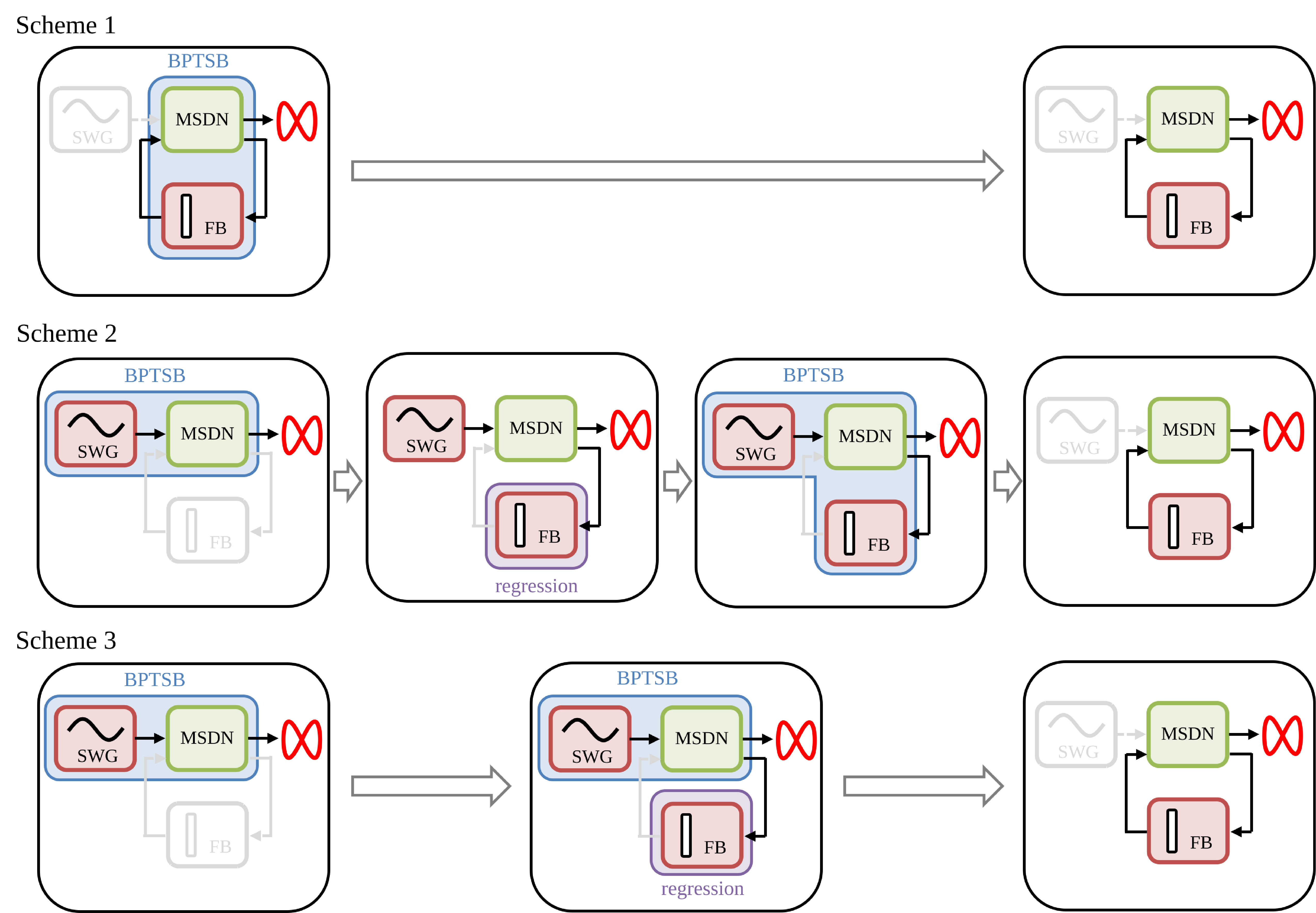}
  \caption{Diagrams representing simultaneous training for a closed loop.
    In Scheme 1, training is performed from the beginning with the MSDN and FL are closed.
    In Scheme 2, BPTSB is applied to the whole system, including the FL after preliminary training with an SWG.
    In Scheme 3, while the SWG and the body are trained by BPTSB, FL is optimized by regression.}
    \label{fig:supp_BPTSB_for_closed-loop}
\end{figure*}

In Section~\ref{sec:CL}, closed-loop control through sensorimotor coupling was realized through functional transfer of the brain.
However, the weights of the feedback layer (FL) were trained using regression, which did not involve simultaneous optimization with the body.
If the FL and the body parameters could be optimized at the same time, it may be possible to achieve more precise closed-loop control.
For such an optimization, we verified three schemes (Fig.~\ref{fig:supp_BPTSB_for_closed-loop}) in the Lissajous curve drawing task.
Note that preliminary training refers to the training via BPTSB for the brain (SWG) and the body to be able to draw one Lissajous curve.
\begin{enumerate}
  \item Without SWG, a closed-loop system including an FL and a body was trained using BPTSB.
  \item After the preliminary training, the SWG, the body, and the FL (still open-loop) were optimized using BPTSB, and then the output of the FL was assigned to the input to the body (closed-loop).
  \item After the preliminary training, the SWG and the body were optimized with BPTSB, and the FL was optimized with regression every time BPTSB updated the parameters. This procedure was repeated many times and a closed loop was formed.
\end{enumerate}
Especially in the second scheme, not only the drawing trajectory error but also the error between the control signals generated in the SWG and the FL were incorporated into this training framework as a loss function as follows:
\begin{align}
  \mathcal{L}=\mathcal{L}_{\text{traj}}+\alpha \mathcal{L}_{\text{FB}},
\end{align}
where $\mathcal{L}_{\text{traj}}$ is the error between the actual trajectory of the CMP and the target trajectory, $\mathcal{L}_{\text{FB}}$ is the error within the outputs of the SWG and the FL, and $\alpha$ is a ratio.
We set $\alpha =0.1$.
The regression was performed only once in order to determine the initial values of the FL as a preliminary step to the BPTSB in Scheme 2.

\begin{table}[t]
  \caption{The results of simultaneous training for a closed loop.
  Experiments were conducted with five different random seeds for three Lissajous curves.
  The table represents the ratio of drawing errors between a closed-loop system with simultaneous training and a closed-loop system without it. 
  Cases in which the error was reduced through simultaneous training are written in bold.}
  \label{table:training_for_replacement}
  \centering
  \subcaption{Scheme 2}
  \begin{tabular}{cccccc}
    \hline
    & 0 & 1 & 2 & 3 & 4  \\ \hline
    $(\alpha,\beta,\delta)=(1,2,0)$ & 1.00 & 1.15 & \textbf{0.87} & 1.09 & \textbf{0.59} \\
    $(\alpha,\beta,\delta)=(2,1,\pi/2)$ & 2.11 & 1.24 & \textbf{0.99} & \textbf{0.68} & 1.04 \\
    $(\alpha,\beta,\delta)=(1,3,0)$ & \textbf{0.97} & 1.26 & \textbf{0.94} & 1.18 & \textbf{0.82} \\
    \hline
  \end{tabular}
  \subcaption{Scheme 3}
  \begin{tabular}{cccccc}
    \hline
    & 0 & 1 & 2 & 3 & 4  \\ \hline
    $(\alpha,\beta,\delta)=(1,2,0)$ & \textbf{0.83} & 1.68 & 1.11 & 1.89 & 1.62 \\
    $(\alpha,\beta,\delta)=(2,1,\pi/2)$ & \textbf{0.58} & 1.56 & \textbf{0.34} & \textbf{0.67} & 1.33 \\
    $(\alpha,\beta,\delta)=(1,3,0)$ & \textbf{0.65} & 1.03 & 1.33 & 1.08 & \textbf{0.66} \\
    \hline
  \end{tabular}
\end{table}

We trained 15 systems and Scheme 1 could not make any systems draw the desired trajectory, even after repeated parameter updates.
All the systems transitioned into a stationary state quickly after the body started moving due to the distribution of the initial values of the feedback weights.
In regard to the second and third schemes, we compared the error before and after applying them.
Only 7 systems out of 15 were improved in terms of the error with Scheme 2, and Scheme 3 enhanced the trajectory accuracy of a mere 6 systems (Table~\ref{table:training_for_replacement}).
Since there were many hyperparameters, including learning rates, in these schemes, it was implied that more sophisticated methods are needed for simultaneous optimization of an FL that maximizes the effective use of the information processing capability of the body.

\subsection{Learning rates}
\label{sec:learning_rates}
The learning rates used for BPTSB are presented in Table~\ref{table:learning-rates}.

\begin{table}[htbp]
  \caption{Learning rates}
  \label{table:learning-rates}
  \centering
  \subcaption{Body parameters}
  \begin{tabular}{cccc}
    \hline
    & Spring constant & Damping coefficient & Rest length \\ \hline
    MNIST & $1\times10^{4}$ & $1\times10^{3}$ & $1\times10^{0}$ \\ 
    Time-series emulation & $1\times10^{4}$ & $1\times10^{3}$ & $1\times10^{1}$ \\
    Lissajous curve (single) & $1\times10^{4}$ & $1\times10^{3}$ & $1\times10^{1}$ \\ 
    Lissajous curves (switching) & $1\times10^{3}$ & $1\times10^{2}$ & $1\times10^{0}$ \\
    Lissajous curves (closed-loop) & $1\times10^{4}$ & $1\times10^{3}$ & $1\times10^{0}$ \\
    Locomotion (forward motion) & $1\times10^{2}$ & $1\times10^{0}$ & $1\times10^{-1}$ \\
    Locomotion (switching) & $1\times10^{0}$ & $1\times10^{-2}$ & $1\times10^{-3}$ \\
    \hline
  \end{tabular}
  \subcaption{Brain parameters}
  \begin{tabular}{cccccc}
    \hline
    & LIL & MLP & Amplitude & Phase \\ \hline
    MNIST & $1\times10^{0}$ & $1\times10^{0}$ & / & / \\ 
    Ttime-series emulation & $1\times10^{-1}$ & $1\times10^{-1}$ & / & / \\
    Lissajous curve (single) & / & / & $1\times10^{1}$ & $1\times10^{2}$ \\ 
    Lissajous curves (switching) & / & / & $1\times10^{0}$ & $1\times10^{1}$ \\
    Lissajous curves (closed-loop) & / & / & $1\times10^{0}$ & $1\times10^{1}$ \\
    Locomotion (forward motion) & / & / & $1\times10^{-3}$ & $1\times10^{-2}$ \\
    Locomotion (switching) & / & / & $1\times10^{-5}$ & $1\times10^{-4}$ \\
    \hline
  \end{tabular}
\end{table}



%

\nocite{*}

\bibliography{thesis}


\section{Acknowledgements}
K. N. is supported by JSPS KAKENHI Grant Numbers 21KK0182 and 23K18472 and by JST CREST Grant Number JPMJCR2014.

\end{document}